\documentclass[11pt]{article}
\pdfoutput=1
\usepackage[preprint]{acl}
\usepackage{times}
\usepackage{latexsym}
\usepackage[T1]{fontenc}
\usepackage[utf8]{inputenc}
\usepackage{microtype}
\usepackage{inconsolata}
\usepackage{graphicx}
\usepackage{inconsolata}
\usepackage{soulutf8}
\usepackage{booktabs}
\usepackage{colortbl}
\usepackage{arydshln}
\usepackage{rotating}
\usepackage{amsmath} 
\usepackage{subcaption}
\usepackage{caption}
\usepackage{placeins}
\usepackage{xspace}
\usepackage{multirow}
\usepackage{enumitem}
\usepackage{authblk} 
\usepackage{fontawesome5} 
\usepackage{epigraph}
\DeclareRobustCommand{\kbname}{\texttt{CHIMERA}\xspace}
\DeclareRobustCommand{\eTitle}{\kbname: A Knowledge Base of Scientific Idea Recombinations for Research Analysis and Ideation}
\definecolor{lightblue}{HTML}{dce8f8}
\definecolor{lightgreen}{HTML}{e1f9e1}
\definecolor{lightred}{HTML}{f9e1e1}
\definecolor{peach}{HTML}{f5e2d5}
\definecolor{mygreen}{HTML}{1F6650}
\definecolor{myred}{HTML}{EA5E5E}

\DeclareRobustCommand{\marker}[1]{\sethlcolor{peach}\hl{#1}}
\DeclareRobustCommand{\bmarker}[1]{\sethlcolor{lightblue}\hl{#1}}

\DeclareRobustCommand{\gr}[1]{\textcolor{mygreen}{\textbf{#1}}}
\DeclareRobustCommand{\rd}[1]{\textcolor{myred}{\textbf{#1}}}

\usepackage[most]{tcolorbox}
\newif\ifcolorversion

\newif\ifshowrev     
\newif\ifshowcr

\showrevfalse
\showcrfalse

\definecolor{revisioncolor}{rgb}{0.2,0.3,0.6}
\definecolor{camerareadycolor}{rgb}{0.1,0.8,0.1}

\DeclareRobustCommand{\revision}[1]{%
  \ifshowrev\textcolor{revisioncolor}{#1}\else#1\fi}

\DeclareRobustCommand{\camera}[1]{%
  \ifshowcr\textcolor{camerareadycolor}{#1}\else#1\fi}

\title{\eTitle}

\author[1]{\textbf{Noy Sternlicht}}
\author[1,2]{\textbf{Tom Hope}}

\affil[1]{\small School of Computer Science and Engineering, The Hebrew University of Jerusalem}
\affil[2]{\small The Allen Institute for AI (AI2)}
\affil[ ]{\small \faGlobe\xspace\href{https://noy-sternlicht.github.io/CHIMERA-Web}{Project}\xspace\xspace\xspace\faGithub\xspace\href{https://github.com/noy-sternlicht/CHIMERA-KB}{Github}\xspace\xspace\xspace\xspace\faDatabase\xspace\href{https://github.com/noy-sternlicht/CHIMERA-KB/tree/main/data}{Data}}

\begin{document}
\maketitle
\begin{abstract}
A hallmark of human innovation is \emph{recombination}---the creation of novel ideas by integrating elements from existing concepts and mechanisms. In this work, we introduce \kbname, the first large-scale Knowledge Base (KB) of recombination examples automatically mined from the scientific literature. \kbname enables empirical analysis of how scientists recombine concepts and draw inspiration from different areas, and enables training models that propose cross-disciplinary research directions. 
To construct this KB, we define a new information extraction task: identifying recombination instances in papers. We curate an expert-annotated dataset and use it to fine-tune an LLM-based extraction model, which we apply to a broad corpus of AI papers. We also demonstrate generalization to a biological domain. We showcase the utility of \kbname through two applications. First, we analyze patterns of recombination across AI subfields. Second, we train a scientific hypothesis generation model using the KB, showing that it can propose directions that researchers rate as inspiring. 
\end{abstract}

\maketitle
\begin{figure}[!htb]
    \centering
    \begin{subfigure}[b]{\columnwidth}
        \centering
        \includegraphics[width=0.85\columnwidth]{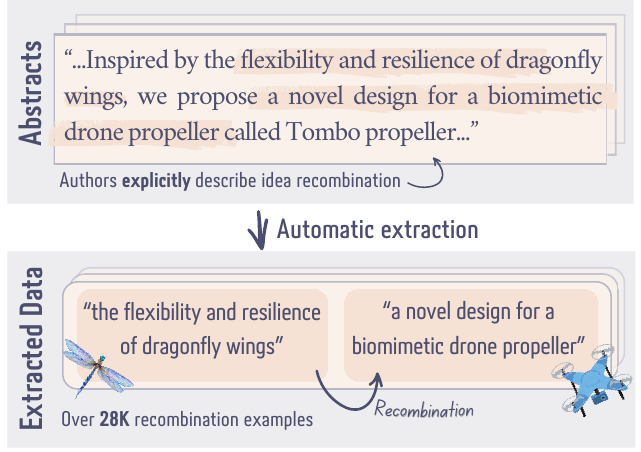}
        \caption{\textbf{Automatic extraction} of recombination examples.}
        \label{fig:fig1a}
    \end{subfigure}
        
    \begin{subfigure}[b]{\columnwidth}
        \centering
        \includegraphics[width=0.85\columnwidth]{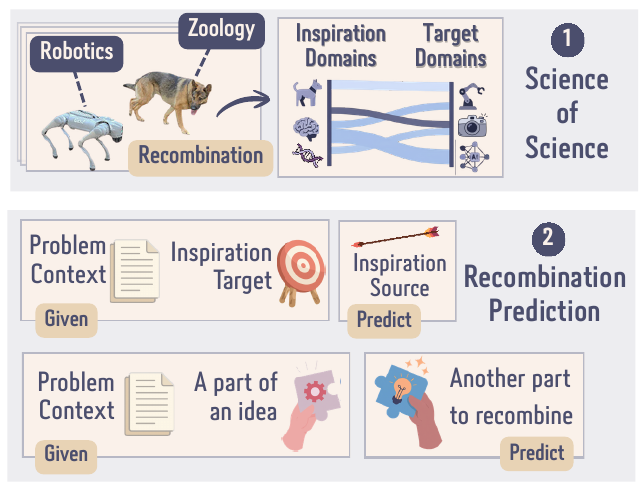}
        \caption{\textbf{Applications} of extracted recombinations.}
        \label{fig:fig1b}
    \end{subfigure}
    \caption{We propose a new task of extracting \emph{recombinations}: examples of how scientists connect ideas in novel ways (\textbf{a}). This data enables applications in research analysis and automated ideation (\textbf{b}).}
    \label{fig:recombination_extraction_example}
    \label{fig:fig1}
\end{figure}

\section{Introduction}\label{sec:introduction}
\epigraph{So much innovation and so many inventions consist of recombination.}{\textit{Joel Mokyr, Economic Sciences Nobel Laureate}}
\textit{Recombination} is a widely recognized mechanism of ideation and innovation \cite{uzzi2013atypical,shi2023surprising}. It involves connecting, reconfiguring, adapting and blending existing concepts, technologies, and resources in novel ways \cite{knoblichConstraintRelaxationChunk1999, mccaffreyInnovationReliesObscure2012}. This often requires forming abstract structural mappings across domains \cite{gentner1997analogy,chan2011benefits}—e.g., as in computational algorithms that take inspiration from nature.

{For decades, researchers have worked on computationally identifying and supporting recombination: from classic work \cite{gentner1983structure,falkenhainer1989structure}, to more modern mixed-initiative systems in the scientific domain \cite{chan2018solvent,kang2022augmenting,choi2024creativeconnect,radensky2024scideator}. At the same time, {science of science} and economics researchers have shown that surprising combinations of ideas often produce high impact \cite{shi2023surprising}. And yet, to date, there does not exist a large-scale repository of naturally occurring recombinations in science. Such a knowledge base could enable training \emph{supervised} recombination models that generate creative directions, evaluating recombination-generation systems, exploring existing recombinations within a given field, and conducting meta-scientific analyses of recombination at scale. 
}

{In this work, we make progress toward this vision}. We first introduce a new task: extracting recombinations from scientific papers. We present \kbname, a large-scale knowledge base (KB) of recombination examples automatically mined from papers. Figure~\ref{fig:fig1a} shows one such case, where a robotic design is inspired by animal mechanics. We then show how \kbname enables analyzing, training and evaluating models on such examples.

Unlike simpler concept co-occurrence methods \cite{Krenn2022ForecastingTF} or general scientific extraction schemas \cite{Luan2018MultiTaskIO}, \kbname targets cases where authors \emph{explicitly} describe recombination as central to their contribution. We focus on two broad recombination types: \emph{\textbf{blends}}, which combine concepts into novel approaches (e.g., augmenting classical ML with quantum computing), and \emph{\textbf{inspirations}}, where ideas from one domain spark solutions in another (e.g., using bird flock behavior to coordinate drones). \kbname captures both intra- and cross-domain cases, including analogies, abstractions, and reductions.

\begin{figure*}[!htb]
\centering
\includegraphics[width=0.95\textwidth]{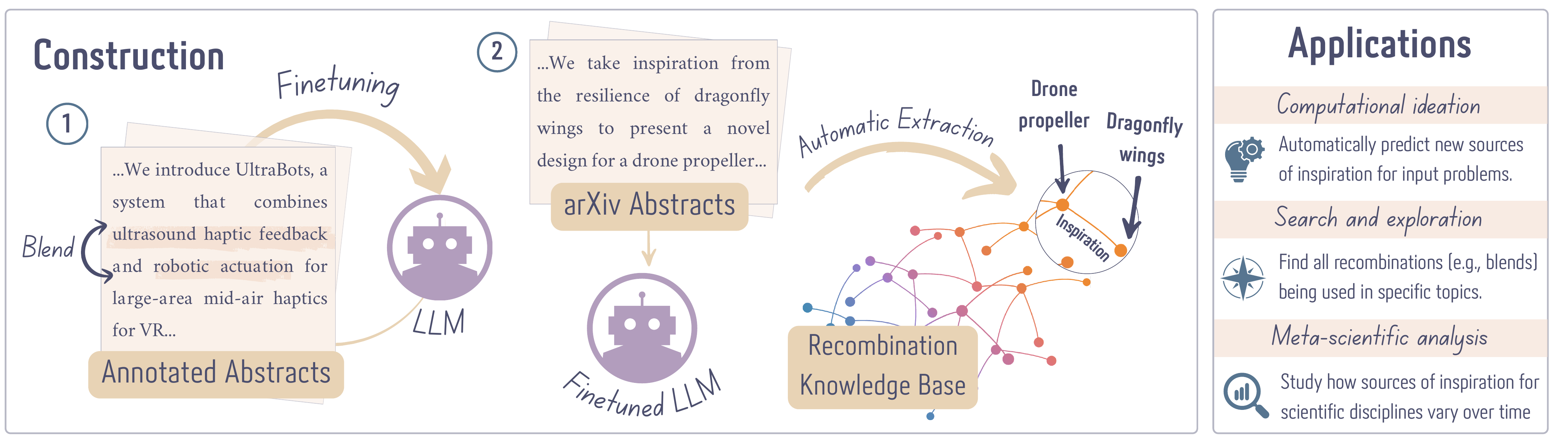}
\caption{\revision{\kbname KB construction and applications. \textbf{Construction}: (1) We use human-annotated recombination examples to fine-tune an LLM for information extraction; (2) the model extracts recombinations from arXiv abstracts to build a large-scale KB. \textbf{Applications}: \kbname supports diverse use cases, including computational ideation, exploration of recombination patterns across scientific domains, and meta-scientific analysis.}}  \label{fig:overview}
\end{figure*}

\revision{The resulting KB and methods enable diverse uses (Figure~\ref{fig:fig1b}, Figure \ref{fig:overview}). We focus on two applications that have seen growing interest in recent years: \textbf{Science Analysis} \cite{Fortunato2018ScienceOS, Wahle2023WeAW, pramanick-etal-2025-nature} and \textbf{Scientific Ideation} \cite{scimon,si2024can, radensky2024scideator, garikaparthi2025irisinteractiveresearchideation}.}

\textbf{Science Analysis.} \revision{We demonstrate how \kbname supports meta-scientific analysis (also known as \textit{science of science}) \cite{Fortunato2018ScienceOS}: empirical studies of how innovation unfolds. 
\kbname enables analysis of how ideas are combined within and across domains \cite{Shi2019SurprisingCO}, and of how disciplines, topics and concepts inspire one another. This provides a \emph{direct and precise} alternative to traditional citation-based \cite{uzzi2013atypical} or co-occurrence-based approaches \cite{Frohnert2024DiscoveringECA}, which are often coarse and noisy. In contrast, \kbname allows to identify how a scientific idea is formed by blending concepts or by taking inspiration from another concept, unlocking new and also more granular analyses.}

\textbf{Scientific Ideation.} \revision{We show how \kbname supports training and evaluating scientific hypothesis generation models \cite{scimon}, by learning from patterns of past recombinations to propose novel concept blends or inspirations (e.g., new analogical inspirations). Prior work has explored suggesting analogical recombinations via unsupervised discovery \cite{radensky2024scideator, hope2017KDD}; in contrast, \kbname provides the first large-scale resource with \emph{real}, author-described examples of how research problems were addressed via recombination. This enables supervised recombination models to observe many examples of how recombinations have been applied to specific problems (e.g., Figure~\ref{fig:fig1}), and \emph{learn} to suggest relevant blends or inspirations for new problems.}

\revision{Finally, \kbname also enables faceted search and exploration \cite{katz2024knowledgenavigatorllmguidedbrowsing}, finding cases of cross-domain inspirations within a topic of interest (e.g., search for all robotics ideas inspired by zoology), sparking new creative directions.}

\revision{To conclude, our contributions are as follows:}
\begin{itemize}[topsep=0pt, itemsep=0pt, parsep=0pt]
\item \revision{We present \kbname, the \textbf{first knowledge base of idea recombination examples} described by authors in scientific papers. \kbname distinguishes between two core types: \emph{blends} and \emph{inspirations}, enabling nuanced analysis in downstream tasks.}

\item \revision{We define a novel extraction task to identify recombinations in scientific abstracts, and release a high-quality, expert-verified dataset of $500+$ manually annotated examples, accompanied by fine-tuned extraction baselines. {Our extraction model, trained on AI papers, further generalizes to a biological domain.}}

\item \revision{We show \kbname's utility through two applications: a) \emph{Meta-scientific analysis} of recombination patterns, and b) \emph{Computational ideation}, where models trained on \kbname propose novel recombination directions.}
\end{itemize}

\section{Related Work}
\paragraph{Recombinant creativity}
\revision{Blending concepts and analogical inspiration are core mechanisms of ideation and innovation in cognitive science and creativity research \cite{McKeown2014TheCS, 2019TheCH, Holyoak1994MentalLA}. These processes involve combining or re-representing existing ideas to produce novel concepts and solutions.}

Recent work explores how idea recombination can enhance LLM-powered ideation tools. For example, \texttt{CreativeConnect} \cite{Choi2023CreativeConnectSR} lets users recombine keywords to generate graphic sketches, while \texttt{Luminate} \cite{Suh2023LuminateSG} supports recombination of dimensional values to produce diverse LLM responses. \texttt{Scideator} \cite{radensky2024scideator} is another recent work that helps researchers explore ideas through interactive concept recombination.
Other studies focus on recombining ideas from input and analogous artifacts \cite{Srinivasan2024ImprovingSO, Chilton2019VisiBlendsAF} \revision{or searching for useful recombinations via iterative idea generation \cite{yang2025moose, yang2025moosechemlargelanguagemodels}}.

In this work, we build \kbname, the first KB of scientific idea recombinations, and show how it enables a new approach for recombinant ideation: training models that \emph{learn} from past examples of how ideas have been recombined in scientific texts, to suggest new recombination directions. 

\paragraph{Scientific IE}
Information extraction (IE) from scientific texts has been widely studied. A foundational resource is \texttt{SciERC} \cite{Luan2018MultiTaskIO}, which labels scientific entities (e.g., methods, tasks, metrics) and generic relations (e.g., conjunction) across $500$ abstracts. Later datasets, such as \texttt{SciREX} \cite{Jain2020SciREXAC} and \texttt{SciDMTAL} \cite{Pan2024SciDMTAL}, expand IE to full documents, but similarly focus on standard schema involving scientific concepts and their relations. \revision{However, existing IE resources are not designed to capture recombinations, often resulting in noisy, irrelevant, or misleading outputs, as illustrated in Appendix~\ref{sec:extraction_comp}, Figure~\ref{fig:extraction_comparison}}.

\revision{In this work, we introduce a focused IE schema tailored specifically to idea recombination, along with a taxonomy that distinguishes between key recombination types: \emph{blend} and \emph{inspiration}. This enables a more precise and semantically rich analysis of cross-domain ideation. For instance, our knowledge base includes numerous analogical inspirations identified in AI research (Figure~\ref{fig:fig1}) - patterns that existing scientific IE schemas fail to capture}.

\begin{table*}[!htb]
    \centering
    \footnotesize
    \begin{tabular}{p{15cm}}
        \toprule
        \multicolumn{1}{c}{\textbf{Automatic Recombination Extraction Examples}} \\
        \midrule
         \textbf{Abstract}: ``...Current archaeology depends on trained experts to carry out bronze dating... we propose  to integrate \sethlcolor{peach}\hl{advanced deep learning techniques} and \sethlcolor{peach}\hl{archaeological knowledge}...'' \newline

         \textbf{Blend}: ``\textit{advanced deep learning techniques}'' $\longleftrightarrow$ ``\textit{archaeological knowledge}'' \\
         \midrule
        \textbf{Abstract}: ``\marker{Click-Through Rate (CTR) prediction} is a pivotal task in product and content recommendation, where \marker{learning effective feature embeddings is of great significance}... inspired by \marker{the Global Workspace Theory in conscious processing}, ... we propose a CTR model that enables Dynamic Embedding Learning with Truncated Conscious...'' \newline

        \textbf{Inspiration}: ``\textit{the Global Workspace Theory...}'' $\longrightarrow$ ``\textit{learning effective feature embeddings for CTR prediction}'' \\
         \bottomrule
    \end{tabular}
    \caption{Example \textbf{blend} and \textbf{inspiration}. Note that blend is a symmetric relation, while inspiration is not.}
    \label{tab:cool_recombination_examples}
\end{table*}
\section{Extracting Recombinations}
\paragraph{Problem definition}\label{methods:problem_definition}
We focus on scientific abstracts where authors \emph{explicitly link} their contribution to a novel combination or clear source of inspiration. 
As outlined in the introduction, we capture this with two coarse-grained relation types: \textbf{blend} and \textbf{inspiration}. \textbf{Blend} refers to the fusion of multiple concepts--such as methods, models, or theories--into a new solution or framework. We use the terms ``concept blend'' and ``concept combination'' interchangeably.
\textbf{Inspiration}, by contrast, refers to transferring knowledge or insight from one entity (the \textit{source}) to another (the \textit{target}). This transfer may be realized through analogies, abstraction, or more general links to influential prior work.

\begin{table}[!htb]
\centering
\footnotesize
\begin{tabular}{@{}lccc@{}}
\toprule
\textbf{Example type} & \textbf{\# Train} & \textbf{\# Test} & \textbf{\# Total}\\
\midrule
\textit{blend} & 124 & 76 &  100 \\
\textit{inspiration} & 45 & 24 & 69 \\
\textit{not-present} & 195 & 116 & 311 \\
\midrule
\textbf{All} & 364 & 216 & 580\\
\bottomrule
\end{tabular}
\caption{Human-annotated corpus. We also include negative examples without recombinations (``\textit{not-present}'').
}\label{table:chimera-s-splits}
\end{table}

Each relation is defined over free-form text spans that represent scientific concepts (see Figure~\ref{fig:fig1}; additional examples in Table~\ref{tab:cool_recombination_examples}). In blend relations, we refer to the participating entities as \textit{combination-elements}; in inspiration relations, we refer to them as the \textit{inspiration-source} and \textit{inspiration-target}.
\revision{This schema captures diverse recombination phenomena, such as metaphor, reduction, or abstraction (as illustrated in Appendix~\ref{sec:fine_grained_types}) while remaining conceptually clear and efficient to annotate. It offers practical annotation advantages and strong alignment with ideation theory \cite{McKeown2014TheCS, 2019TheCH, Holyoak1994MentalLA}}.

\subsection{Recombination Mining}\label{subsection:recombination_mining}
We begin by curating a dataset of annotated recombination examples, which we use to train an information extraction model. The trained model is then applied to extract recombinations at scale. This process is illustrated in Figure~\ref{fig:overview}.

\paragraph{Data sourcing}
We annotate AI-related arXiv papers \cite{Saier2020unarXiveAL}\footnote{We focus on the following \href{https://arxiv.org/category_taxonomy}{arXiv categories}: cs.AI, cs.CL, cs.CV, cs.CY, cs.HC, cs.IR, cs.LG, cs.RO, cs.SI}. The data undergo an initial keyword-based filtering to identify works that are more likely to specify idea recombination (keywords in table \ref{table:keywords}, appendix \ref{sec:recomb_keywords}).

\paragraph{Annotation process}
\revision{Our annotation setup follows standard IE practices, using two trained annotators and expert review to balance quality and feasibility \cite{naik2024care, Sharif2024ExplicitIA, pramanick-etal-2025-nature}.} 
\revision{Following a screening phase, we recruited two annotators with scientific PhDs via Upwork\footnote{\url{https://www.upwork.com}}, selected from a pool of highly experienced workers we had previously collaborated with. Screening involved annotating examples using a detailed guidelines document
\footnote{\url{https://tinyurl.com/zy27uhdp}}
, followed by a one-hour training session covering additional examples and edge cases.} Annotation was conducted using \texttt{LightTag} \cite{Perry2021LightTagTA}, a web-based annotation platform. This process yielded $580$ annotated abstracts, summarized in Table~\ref{table:chimera-s-splits}. To monitor annotation quality, we assign 10\% of the examples to both annotators and review this shared subset after each batch. Disagreements are resolved through discussion and revision. All annotations are then reviewed by an NLP expert, who verifies correctness, refines spans, and consolidates annotations.

\begin{table}[!htb]
\centering
\footnotesize
\begin{tabular}{@{}lcc@{}}
\toprule
\textbf{Category}  & \textbf{\# Interdisciplinary} & \textbf{\# Total}\\
\midrule
Inspiration Edges  & 5,182 (54.1\%) & 9,578 \\
Blend Edges  & 1,792 (9.6\%) & 18,586 \\ \midrule
Edges (total)   & 6,974 (24.8\%) & 28,164 \\
Nodes (total)   & n/a  & 43,393\\ 
\bottomrule
\end{tabular}
\caption{\kbname contains over 28K recombinations, a quarter of them interdisciplinary.}\label{table:kg_summary}
\end{table}

\begin{table*}[!htb]
\centering
\footnotesize
\begin{tabular}{p{3.6cm}|lccc}
\toprule
\textbf{Task}& \textbf{Baseline} & \textbf{Precision}  & \textbf{Recall} & \textbf{F1} \\ 
\midrule
\multirow{7}{3.6cm}{\textit{\textbf{Abstract classification}:\newline\newline Does it discuss a recombination?}} & \cellcolor{peach}Human-agreement & \cellcolor{peach} 0.786 & \cellcolor{peach}0.795 & \cellcolor{peach} 0.789 \\ 
& E2E$_{Mistral-7B-Instruct-v0.3}$ & \textbf{0.815} & \textbf{0.762} & \textbf{0.763} \\
& E2E$_{Llama-3.1-8B-Instruct}$ & 0.630	&	0.628	&	0.620 \\
& E2E$_{GoLLIE-13B}$& 0.677 & 0.667 & 0.667 \\
& E2E$_{GPT-4o}$ &0.720 &0.580 & 0.572 \\
& Abstract-classifier$_{Mistral-7B-Instruct-v0.3}$ & 0.622& 0.607& 0.602 \\
& Abstract-classifier-CoT$_{Mistral-7B-Instruct-v0.3}$& 0.774 & 0.748 &0.749 \\
\midrule
\multirow{8}{3.6cm}{\textit{\textbf{Entity extraction}:\newline\newline What are the relevant entities?}} & \cellcolor{peach}Human-agreement & \cellcolor{peach}0.863 & \cellcolor{peach}0.585 &  \cellcolor{peach}0.665 \\ 
& E2E$_{Mistral-7B-Instruct-v0.3}$  & \textbf{0.587} & 0.352 & \textbf{0.440} \\
& E2E$_{Llama-3.1-8B-Instruct}$ & 0.249	&	0.259	&	0.252 \\
& E2E$_{GoLLIE-13B}$& 0.259 & 0.187 & 0.217 \\
& E2E$_{GPT-4o}$ & 0.138 & 0.293 & 0.217 \\
& Entity-extractor$_{GPT-4o}$ & 0.268 &  0.263 & 0.247 \\
& Entity-extractor$_{SciBERT}$ & 0.324 & 0.248 & 0.276 \\
& Entity-extractor$_{PURE_{SciBERT}}$ & 0.187 & \textbf{0.536} & 0.271 \\
\midrule
\multirow{5}{3.6cm}{\textit{\textbf{Relation extraction}:\newline\newline What is the recombination?}} &\cellcolor{peach}Human-agreement & \cellcolor{peach}0.793 & \cellcolor{peach}0.574 & \cellcolor{peach}0.641\\ 
& E2E$_{Mistral-7B-Instruct-v0.3}$  & \textbf{0.598} & 0.366 & \textbf{0.454}\\
& E2E$_{Llama-3.1-8B-Instruct}$ & 0.264	&	0.294	&	0.276 \\
& E2E$_{GoLLIE-13B}$ & 0.301 & 0.219 & 0.253 \\
& E2E$_{ICL-GPT-4o}$ & 0.223 & \textbf{0.385} & 0.244 \\
\bottomrule
\end{tabular}
\caption{Recombination extraction results with fine-tuned or prompted LLMs. Models approach inter-annotator agreement (IAA) on abstract classification---detecting whether a recombination is present. For entity and relation extraction, there is a large gap between models and IAA. However, our analyses reveal that many extraction errors in \kbname are rather minor (e.g., span boundaries). }
\label{table:main_results_divided}
\end{table*}

\paragraph{Automatic recombination mining} We use the collected data to fine-tune an LLM-based extraction model. We instruct the model to extract the most salient recombination from the text, if one exists. The model must determine whether the text discusses recombination, infer its type, and identify entities in a single query. We devise the test set from examples where at least two annotators (out of three) agree on the recombination type (or absence), ensuring high-quality, low-ambiguity data. Table \ref{table:chimera-s-splits} summarizes the train and test sets.

\subsection{The \kbname Knowledge Base}\label{subsection:kg}
We construct the \kbname knowledge base by mining recombination examples from scientific abstracts, categorizing them, and representing them in a graph where nodes are scientific concepts and edges denote recombination relations.

\paragraph{Large-scale mining} We use abstracts from the arXiv dataset\footnote{\url{https://tinyurl.com/mrzksbky}}, which updates monthly and includes more recent papers than unarXive \cite{Saier2020unarXiveAL}.
We apply our fine-tuned extraction model over publications from $2019$-$2024$ within the same CS categories used for the annotation task. We then filter out predictions that don't conform to the data schema or cannot be parsed. 
\camera{Unlike human annotation, we apply no keyword filtering. The top $30$ keywords in the extracted KB (Appendix~\ref{sec:keyword_freq}) reveal diverse recombination sub-types, spanning inspiration (\textit{``inspired''}), integration (\textit{``fusion''}), structural alignment  (\textit{``align''}, \textit{``view''}), and abstraction (\textit{``cast''}). We also notice many analogy-related keywords. Beyond that, the model detects recombinations in abstracts with no seed keywords, demonstrating generalization to unseen recombination-expressing language (\textit{``we draw on...''}). See examples in Appendix~\ref{sec:keyword_freq}.}

\paragraph{Categorization} We apply \texttt{GPT-4o} to identify the scientific domain of each extracted entity given the abstract. This enables analyses we perform in Section \ref{subsec2:kb_analysis}. Further, each node is assigned a higher-level discipline—either the arXiv group name (e.g., ``\textit{computer-science}'' for cs.AI) or a relevant non-arXiv domain. Additional technical details regarding this step appear in Appendix~\ref{sec:domain_anlysis}.

\paragraph{KB building} We normalize entities by clustering semantically highly similar ones. We enrich each edge in the graph with the publication date and arXiv categories of the paper. For simplicity, we focus on binary relations. Table~\ref{table:kg_summary} summarizes the resulting KB, including counts of interdisciplinary blends and inspirations. {Appendix \ref{sec:entity_norm} provides additional details about entity normalization.}

\paragraph {Post-processing}We further refine extracted entities using a SOTA LLM (\texttt{claude-opus-4.6}). A qualitative review shows improved quality over the original extraction. We performed this process after the initial KB construction, therefore, we added post-processed fields to the dataset alongside the original ones. All reported results in this work are without post-processing, as we used the best models available at the time of running our original experiments. Additional details are in Appendix~\ref{fig:postprocessing_prompt}.

\section{Results}\label{sec:results}

\subsection{Experimental Settings}\label{eval_criteria}
\paragraph{Evaluation criteria} We evaluate (1) \textbf{Abstract classification}--does the text discuss recombination? (2) \textbf{Entity extraction}--what entities are described? and (3) \textbf{Relation extraction}--what is the relation discussed?
For abstract classification, we report precision, recall, and F1.
For entity and relation extraction, we adopt a soft matching approach: two entities of the same type match if they refer to semantically similar concepts. We use \texttt{GPT-4o-mini}\footnote{Performed on par with \texttt{GPT-4o}.} to judge similarity (see prompt and details in Appendix~\ref{sec:span_sim_prompt}).

\begin{table}[!htb]
    \centering
    \footnotesize
    \begin{tabular}{@{}p{7.5cm}@{}}
        \toprule

        \textbf{Abstract}: ``Living cells inherently exhibit the ability to spontaneously reorganize their structures in response to changes... Among these responses, \sethlcolor{peach}\hl{the organization of stress fibers composed of actin molecules changes in direct accordance with the mechanical stiffness of their environments}... These transformations share similarities with \sethlcolor{peach}\hl{phase transitions studied in condensed matter physics}, yet despite extensive research on cellular dynamics, the introduction of the statistical mechanics perspective to the environmental dependence of intracellular structures remains underexplored...'' \newline

        \textbf{Inspiration}: ``\textit{phase transitions studied in condensed matter physics}'' $\longrightarrow$ ``\textit{the organization of stress fibers...}''. \\

         \bottomrule
    \end{tabular}
       \caption{{Our extraction model generalizes beyond the domain it was trained on, maintaining high accuracy.}}
    \label{tab:bio-recomb-examples}
\end{table}

A predicted entity may match at most one gold entity, and vice versa; extra matches are ignored. We compute precision, recall, and F1 under this soft matching. For relations, we use partial matching: a predicted relation contributes to the true positive count proportionally to the number of correctly matched entities in a gold relation of the same type. {We measure inter-annotator agreement using the same precision, recall, and F1, following standard practice in IE, where one annotator is treated as the gold \cite{naik2024care, Sharif2024ExplicitIA}.}

\paragraph{Extraction baselines}
We evaluate several extraction baselines available at the time of conducting our experiments, including end-to-end (E2E) models that jointly predict whether an abstract discusses recombination, identify its type, and extract the involved entities (prediction of any relation indicates positive abstract classification). We also train sub-task models: \emph{abstract classifiers}, which detect whether recombination is discussed, and \emph{entity extractors}. Implementation details are in Appendix~\ref{sec:extraction_baselines}. \revision{To contextualize model performance, we compare results against inter-annotator agreement, used as a proxy for human-level performance. Appendix~\ref{sec:IAA} presents additional details concerning agreement computation}.

 \begin{table}[!htb]
    \centering
    \footnotesize
    \begin{tabular}{@{}p{7.5cm}@{}}
        \toprule
        \textbf{Abstract}: ``Efficient exploration of large-scale environments remains a critical challenge in robotics... The presented bio-inspired framework heuristically models \marker{frontier exploration} similar to the \marker{shepherding behavior of herding dogs}. This is achieved by modeling frontiers as a sheep swarm reacting to robots modeled as shepherding dogs...'' \newline

        \textbf{Inspiration}: ``\textit{the shepherding behavior of herding dogs}'' [\bmarker{zoology}] $\longrightarrow$ ``\textit{Frontier exploration}'' [\bmarker{cs.ro}] \\
        \midrule
        \textbf{Abstract}: ``\marker{Histopathological image classification} constitutes a pivotal task in computer-aided diagnostics...Inspired by \marker{the multi-granular diagnostic approach of pathologists}, we perform feature extraction on cell structures at coarse, medium, and fine granularity, enabling the model to fully harness the information in histopathological images...'' \newline

        \textbf{Inspiration}: ``\textit{the multi-granular diagnostic approach of pathologists}'' [\bmarker{biomedical sciences}] $\rightarrow$ ``\textit{Histopathological image classification}'' [\bmarker{cs.cv}] \\
        \bottomrule
    \end{tabular}
    \caption{Extracted \bmarker{\textit{inter-domain}} inspiration examples.}\label{table:inspiration_examples}
\end{table}

\begin{figure*}[!htb]
    \centering
    
    \begin{subfigure}[b]{\columnwidth}
        \centering
        \includegraphics[width=0.95\textwidth]{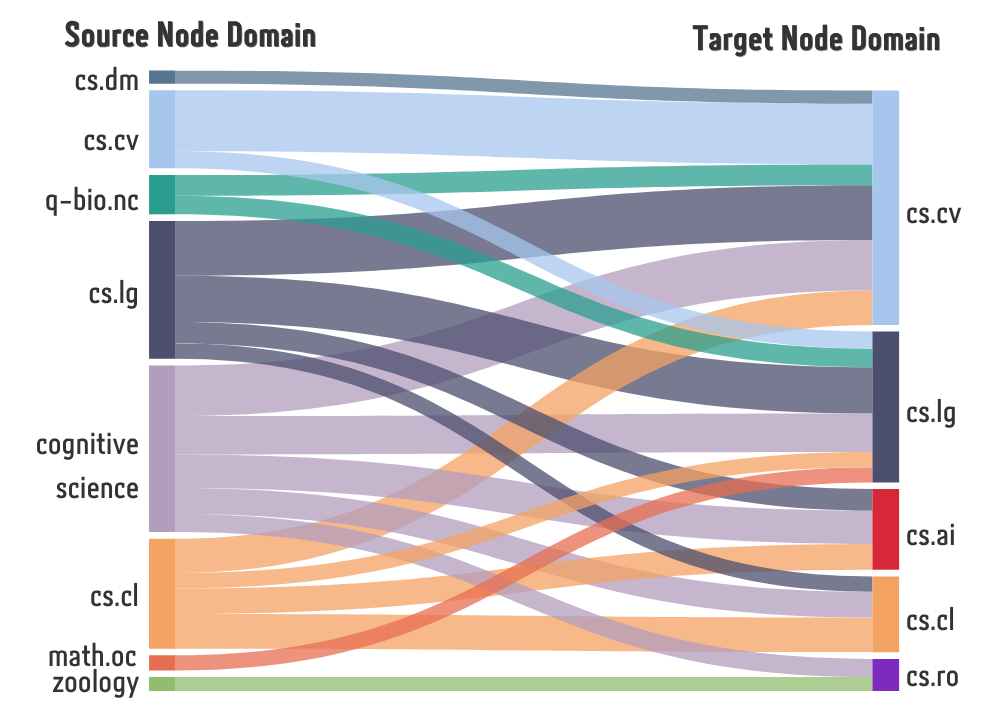}
        \caption{\revision{Frequent domains in \textbf{inspiration} edges.}}
        \label{fig:inspiration_sunkey}
    \end{subfigure}
    \hspace{0.0\columnwidth}
    \begin{subfigure}[b]{1.05\columnwidth}
        \centering
        \includegraphics[width=\textwidth]{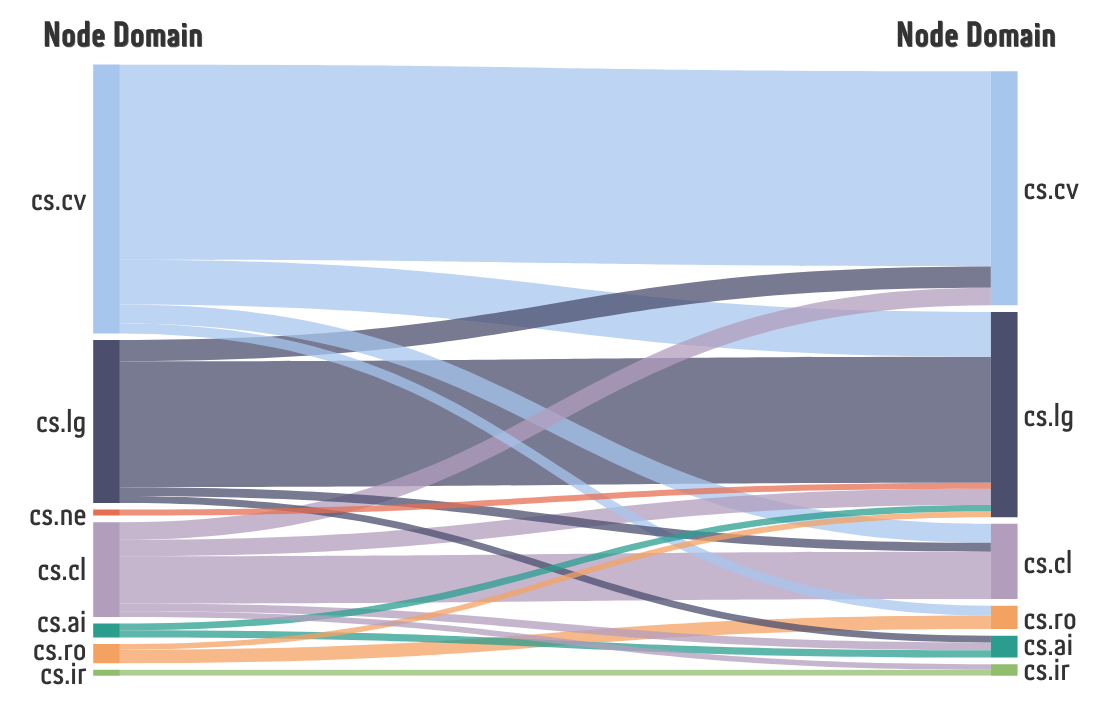}
        \caption{\revision{Frequent domains in \textbf{blend} edges.}}
        \label{fig:bleds_sunkey}
    \end{subfigure}


    \begin{subfigure}[b]{\textwidth}
        \centering
        \includegraphics[width=0.97\textwidth]{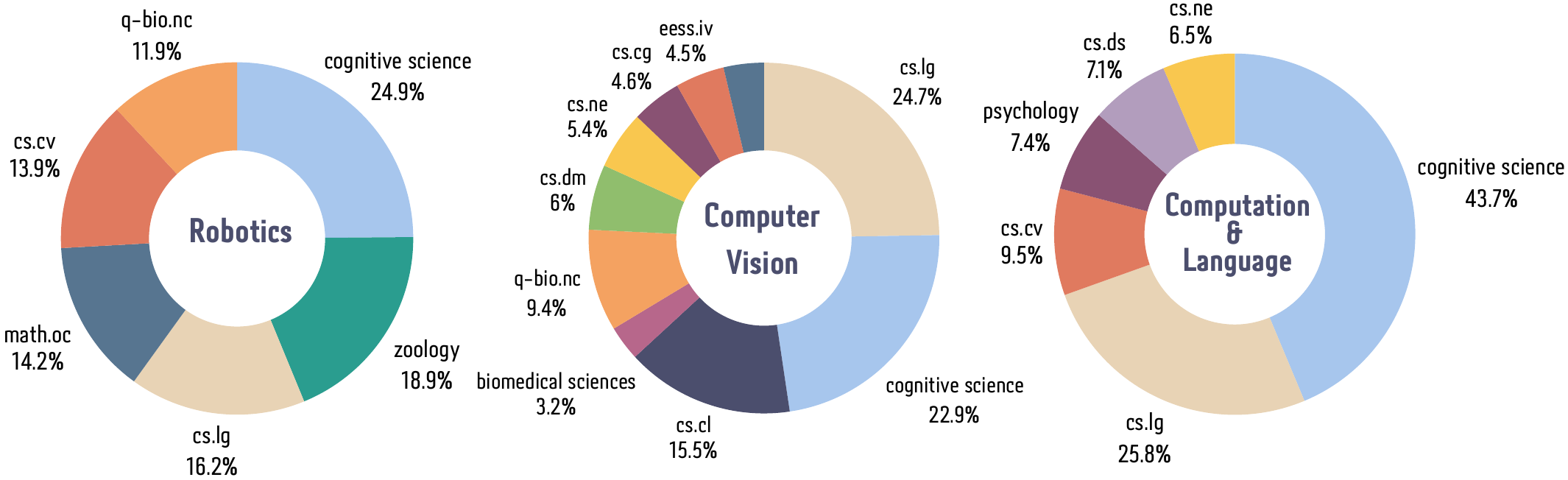}
        \caption{\revision{Common \textbf{sources of inspiration} in leading domains.}}
        \label{fig:domain_analysis}
    \end{subfigure}
    
    \caption{Recombinations between areas. \textit{cs.*}, \textit{q-bio.nc} and \textit{math.oc} are arXiv categories. Inspirational connections are often cross-domain (Figure \ref{fig:inspiration_sunkey}), whereas blends tend to occur within the same domain (Figure \ref{fig:bleds_sunkey}). Figure \ref{fig:domain_analysis} zooms in on a few domains, for example, revealing that \textit{robotics} often draws inspiration from \textit{zoology}.}
    \label{fig:frequent_recomb_sunkey}
\end{figure*}

\subsection{Extraction Results}\label{subsec:extraction_res}
Table \ref{table:main_results_divided} reports results for abstract classification, entity extraction, and relation extraction. Human agreement scores are $0.760$, $0.675$, and $0.651$ respectively, aligning with soft annotator agreement reported in similar complex extraction tasks \cite{naik2024care, Sharif2024ExplicitIA}. \revision{Cohen’s $\kappa$ also indicates moderate to substantial agreement: $\kappa = 0.578$ for abstract classification, $0.631$ for entity extraction, and $0.542$ for relation extraction. Analysis of annotator disagreement appears in Appendix~\ref{subsec:disagreement}---most disagreements concern the presence of a recombination or span boundaries, while disagreements over recombination type are rare.}

Fine-tuning \texttt{Mistral-7B} on our data yields the best performance across all subtasks. We observe that entity and relation extraction are more challenging than classification for both humans and SOTA LLMs. However, humans still significantly outperform automatic extraction approaches. Appendix~\ref{sec:qualitative_analysis} presents an analysis of extraction errors. Interestingly, focusing on a smaller portion of the recombination extraction task is not necessarily easier than performing it end-to-end, as seen in the lower performance of abstract classifiers. We discuss this point further in Appendix \ref{sec:e2e_vs_spacial}.

\paragraph{Large-scale evaluation}\label{par:extraction_accuracy}
\revision{To assess extraction quality at scale, we evaluate $2,000$ \kbname examples using an LLM judge (\texttt{GPT-4.1}). An example is labeled correct if (1) extracted entities reflect meaningful scientific concepts, and (2) their relation captures a key recombination explicitly described in the abstract. We validate the judge’s reliability by showing high agreement with human annotations. Applied to the full sample, the judge estimates an extraction accuracy of $80.55$\%, supporting the robustness of our approach. Notably, most extraction errors are minor, typically involving correct recombinations with extracted entities less informative than those in the original abstract, suggesting the judge’s estimate is conservative (examples and additional details in Appendix~\ref{section:qualitative_extraction_eval})}. 

\begin{figure}[!htb]
    \centering
    \includegraphics[width=0.9\columnwidth]{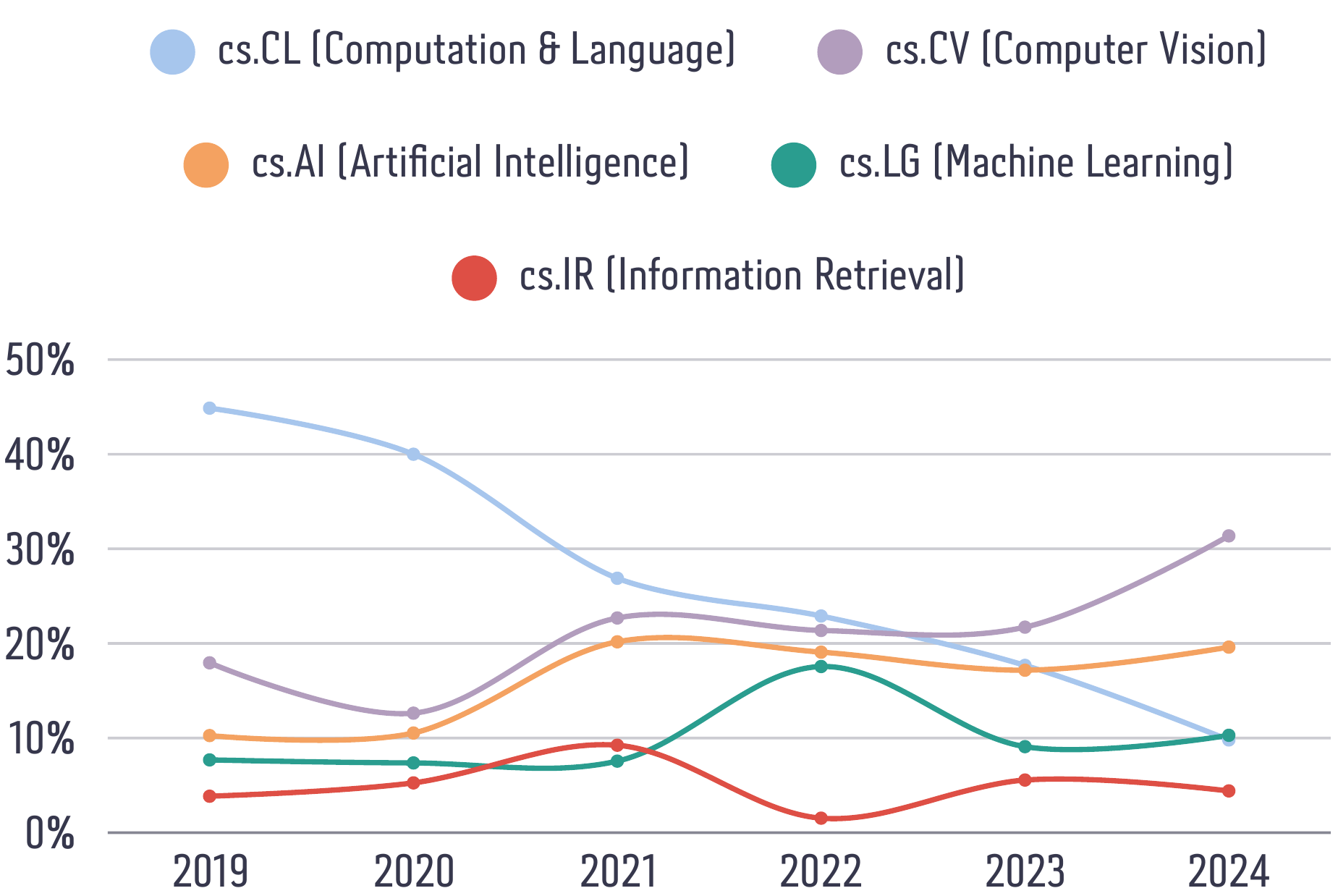}
    \caption{Prevalent domains inspired by \textit{cs.CL} concepts (NLP). Note the \emph{decrease} in within-domain inspiration.}
    \label{fig:nlp_inspiring_other_domains}
\end{figure}

\begin{figure*}
    \centering
    \includegraphics[width=0.9\textwidth]{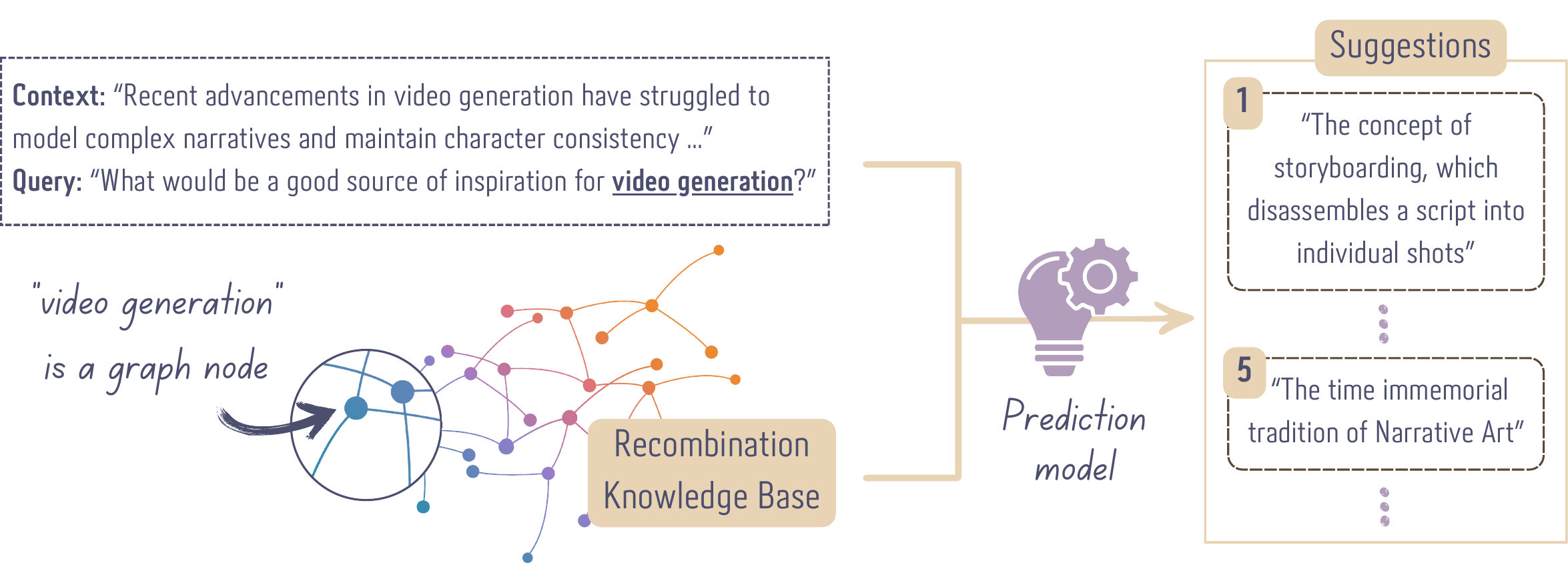}
    \caption{\revision{Recombination prediction. Given a context string and a query about recombining a graph node, a model trained on \kbname suggests plausible recombination directions, leveraging patterns \emph{learned} from prior examples.}}
    \label{fig:prediction_overview}
\end{figure*}

\paragraph{Domain generalization}\label{par:generalization}
{We evaluated the model's ability to generalize to out-of-distribution domains by extracting recombinations from 100 quantitative biology abstracts (arXiv:q-bio). Despite being trained solely on AI-related categories (as discussed in Section \ref{subsection:recombination_mining}), our model generalizes well, achieving 94\% accuracy in our automatic LLM-as-a-Judge evaluation setup. Table \ref{tab:bio-recomb-examples} provides a representative extraction example.}

\subsection{KB Meta-Science Analysis}\label{subsec2:kb_analysis}

\paragraph{Blends vs. inspirations} Figures \ref{fig:inspiration_sunkey} and \ref{fig:bleds_sunkey} present the predominant domain pairs for inspiration and blend relations in \kbname (above the $0.9$ quantile). The analysis reveals an interesting pattern of a distinct difference in behavior between inspirations and blends: inspirations span a broader range of domains, while blends tend to link within the same or similar domains. This suggests that when human researchers take inspiration they tend to look across more areas other than their own, but tend to look within their own domain when they build approaches by integrating together mechanisms. Inspirations also link more often to areas not covered by the arXiv taxonomy, e.g., cognitive science and zoology. More research building on our initial analysis and KB can shed additional light on the different ways in which scientists combine concepts to form ideas. Table \ref{table:all_inspirations} in Appendix \ref{sec:predominant_recomb} provides a tabular view of this analysis for clarity.

\begin{table}[!htb]
\centering
\footnotesize
\begin{tabular}{@{}lccc@{}}
\toprule
\textbf{Split} & \textbf{\# Inspiration} & \textbf{\# Blend} & \textbf{\# Total}\\
\midrule
Train & 5,408 & 19,909&  25,317\\
Validation& 119 & 411 & 530 \\
Test &  2,026 & 8,591 & 10,617\\
\bottomrule
\end{tabular}
\caption{We split prediction data by the publication years associated with each query (training and validation sets $<$ $2024$, test set $\geq$ $2024$) to avoid contamination.}\label{table:pred_data_splits}
\end{table}

\paragraph{Inspiration analysis} We next analyze how different fields draw inspiration from each other. Figure \ref{fig:domain_analysis} shows the top $10$\% cross-domain inspiration sources for three prevalent domains in the graph: \textbf{cs.RO} (Robotics), \textbf{cs.CV} (Computer Vision) and \textbf{cs.CL} (NLP). We observe that while some sources of inspiration (like \textit{cognitive-science}) are commonly shared across related fields, domains may draw inspiration from unique sources (e.g., from \textit{zoology} to \textit{cs.RO} as seen in Figure \ref{fig:fig1}). Interestingly, {cs.CV} takes more inspiration from {cs.CL} than vice versa. {cs.CL} also takes considerably more inspiration from cognitive science than {cs.CV}, and also takes inspiration from psychology (see example in Table \ref{tab:cool_recombination_examples}), while {cs.CV} takes more inspiration from biomedical sources. {cs.CV} also takes inspiration from mathematical topics (discrete math, optimization and control). Table \ref{table:inspiration_examples} presents examples of such interdisciplinary inspirations.

Figure \ref{fig:nlp_inspiring_other_domains} shows the percentage of target nodes in domains drawing inspiration from cs.CL (NLP) over time. We observe a decrease in intra-domain inspiration (cs.CL-cs.CL inspirations), and an increase in cs.CL inspiring cs.CV (Computer Vision).

\begin{table*}[!htb]
\centering
\footnotesize
\begin{tabular}{@{}p{5.1cm}ccccccc@{}}
\toprule
\textbf{Baseline} & \faIcon{arrow-up}\textbf{H@3} & \faIcon{arrow-up}\textbf{H@5} & \faIcon{arrow-up}\textbf{H@10} & \faIcon{arrow-up}\textbf{H@50} & \faIcon{arrow-up}\textbf{H@100} & \faIcon{arrow-up}\textbf{MRR} & \faIcon{arrow-down}\textbf{MedR}\\
\midrule
all-mpnet-base-v2  & 0.033 & 0.042 & 0.061 & 0.126 & 0.170 & 0.033 & 1305 \\
bge-large-en-v1.5 & 0.041 & 0.053 & 0.076 & 0.151 & 0.199 & 0.041 & 1135 \\
e5-large-v2 & 0.024 & 0.033 & 0.050 & 0.113 & 0.155 & 0.026 & 1590 \\
\hline
all-mpnet-base-v2$_{finetuned}$ & \textbf{0.110} & \textbf{0.135} & 0.178 & \textbf{0.320} & \textbf{0.402} & \textbf{0.106} & \textbf{194} \\
bge-large-en-v1.5$_{finetuned}$ & 0.104 & 0.130 & 0.168 & 0.306 & 0.392 & 0.102 & 222 \\
e5-large-v2$_{finetuned}$& 0.107 & 0.133 & 0.173 & 0.317 & 0.397 & 0.103 & 212\\
\hline
all-mpnet-base-v2$_{finetuned}$ + RankGPT & 0.100 & 0.130 & \textbf{0.192} & \textbf{0.320} & \textbf{0.402} & 0.097 & \textbf{194}\\
\bottomrule
\end{tabular}
\caption{Recombination prediction results. \textbf{MedR} = Median Rank. Fine-tuning on \kbname improves MedR \textbf{10×}. Interestingly, reranking the top-$20$ answers using RankGPT boosts the H@10 but slightly reduces H@3,5 and MRR.}\label{table:recomb_pred_results}%
\end{table*}

\section{Recombination Prediction}
\label{subsection:recomb_pred}
We demonstrate how \kbname could be used to train supervised models that recombine concepts and suggest scientific ideas. While \kbname could be used for training generative models, in this work we begin with models in a \emph{contextualized} link prediction setting, which is important in literature-based discovery \cite{sosa2022contexts, scimon}. \revision{Figure \ref{fig:prediction_overview} illustrates the recombination prediction task. Our data is a mixture of two variations on recombination prediction, corresponding to predicting  \emph{inspirations} and predicting \emph{blends}: 

\begin{enumerate}[nosep]
    \item Given a problem context and a \emph{target} (a purpose, goal), predict a \emph{source} of inspiration.
    \item Given a problem context and a \emph{part} of the idea, predict another part to blend. 
\end{enumerate}
Specifically, given a natural language context (e.g., ``\textit{Advancements in video generation have struggled to model complex narratives...}'') and a query about recombining a graph node (e.g., ``\textit{What would be a good source of \textbf{inspiration} for \textbf{video generation}?}'') the goal is to predict an entity to complete the recombination (e.g., ``\textit{The concept of storyboarding...}''). Formally, given a query with context (e.g., a problem, experimental settings), an entity $e$ and a recombination type $\tau$, the task is to predict an entity $e'$ such that $(e, \tau, e')$ is a valid edge in \kbname.}

\paragraph{Data preparation} We start by converting edges to pairs of queries and answers. The queries describe the task inputs: a single graph node, the edge recombination type, and a context string, which we extract from the corresponding abstract using \texttt{GPT-4o-mini}. Note that this process might leak information regarding the answer (the other graph node) into the query. Therefore, we follow it by applying \texttt{GPT-4o-mini} to identify leakages (see examples and implementation details for this step in Appendix \ref{sec:pred_data_prep}). We discard approximately $22$\% of pairs due to leaks and split the remainder by publication year, with all papers published after 2024 in the test set. Table \ref{table:pred_data_splits} summarizes the data splits.

\begin{figure}[!htb]
    \centering
    \includegraphics[width=0.8\columnwidth]{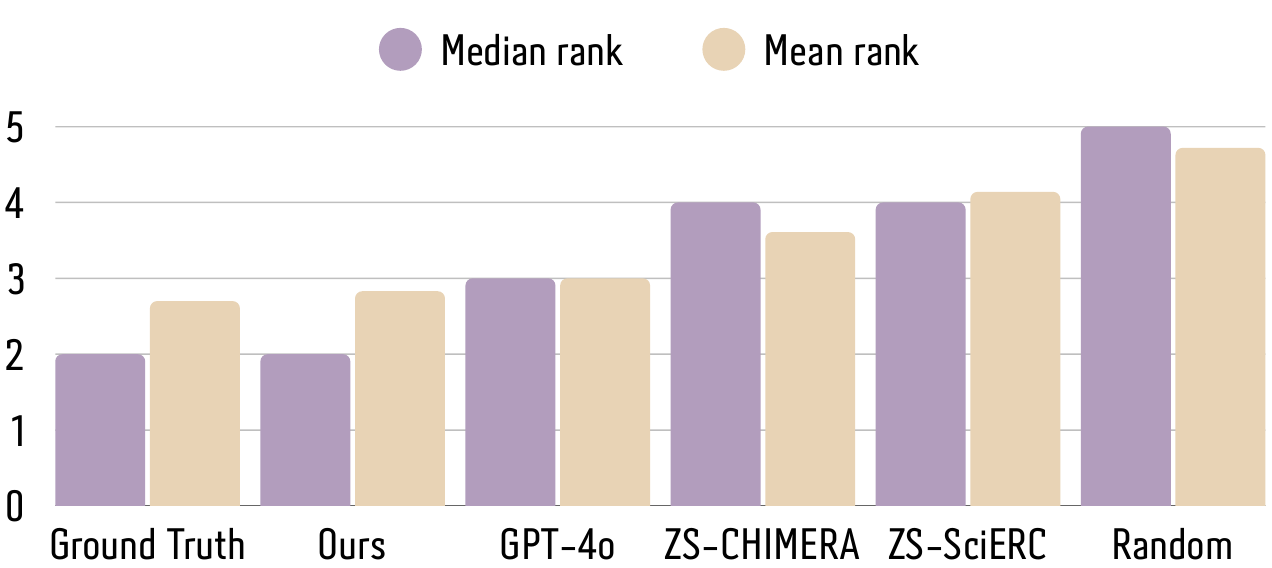}
    \caption{\revision{Researchers find our recombination suggestions almost \textit{as helpful as the gold answer} in inspiring ideas, validating our automated evaluation.}}
    \label{fig:user_study_results}
\end{figure}

\paragraph{Prediction} We experiment with zero-shot and finetuned retrievers based on encoders trained before the test set cutoff year (2024). We next explore applying a \texttt{GPT-4o}-based reranker \cite{Sun2023IsCG} to the top 20 retrieved results to improve our predictions further.  The \texttt{GPT-4o} data cutoff is October $2023$, meaning the reranker is also unfamiliar with our test set. Appendix \ref{sec:prediction_baselines} provides additional implementation details for the prediction baselines.

\subsection{Prediction Results}\label{subsec2:recombination_pred_results}
Table \ref{table:recomb_pred_results} presents our results. Fine-tuning greatly improves retrievers, decreasing the median rank of the gold answer by an order of magnitude. The last row reports results using RankGPT~\cite{Sun2023IsCG} with \texttt{GPT-4o} as a reranker, applied to the top-20 candidates from the best-performing retriever (all-mpnet-base-v2$_\text{finetuned}$). While reranking improves Hits@10, it lowers performance on Hits@3, Hits@5, and MRR. \revision{These seemingly counterintuitive results are further examined in Appendix~\ref{sec:reranker_errors}. We find that the reranker can inadvertently lower the rank of the gold answer in cases where (i) multiple plausible answers are present, or (ii) the gold answer appears alongside semantically similar variants, making it difficult to distinguish between highly relevant alternatives and the annotated gold.}

\paragraph{User study}\label{subsubsection:user_study}
We recruit \revision{five volunteers} with verified research experience (at least one published paper) and assign them examples based on their expertise. Each example includes an inspiration query and suggestions from six sources: (1) \textbf{Ours}: our method (with reranking) (2) \textbf{Ground-Truth}: the gold answer, (3) \textbf{Random}: a random test-set node, (4) \textbf{GPT-4o}: a \texttt{GPT-4o} generated suggestion, (5) \textbf{ZS-\kbname}: zero-shot prediction using our test nodes as candidates, and (6) \textbf{ZS-SciERC}: zero-shot prediction using SciERC-extracted candidates \cite{Luan2018MultiTaskIO}. For baselines returning a ranked list of suggestions, we only use the top one.

Annotators ranked suggestions by their \emph{helpfulness} in inspiring interesting ideas. \revision{Figure \ref{fig:user_study_results} reports the median and average rank across $100$ examples} (lower values are better). Our approach receives a similar rank as the gold answer, and annotators prefer it to all other baselines. This complements the automatic evaluation, showing that we learn to create helpful recombinations. See \revision{Appendix~\ref{app:user_study} for more study details and examples of model predictions that participants found especially inspiring}.

\section*{Conclusion}

We introduced \kbname, the first large-scale knowledge base of idea recombinations in scientific literature. To build \kbname, we formulated a new task of extracting from papers recombinations in the form of \emph{blends} and \emph{inspirations}. \kbname moves beyond coarse keyword co-occurrence or co-citation signals to capture explicit, fine-grained information on how scientists draw inspirations across areas and form links between methods and concepts. We showed that \kbname supports meta-scientific analyses; e.g., we found that while researchers tend to use inspirations from outside their area, they tend to look within their own domain when they integrate together mechanisms. We then used \kbname to train and evaluate supervised ideation models. We formulated a recombination prediction task with two variants: \emph{inspiration} prediction for a given problem, and \emph{blend} prediction to predict a complementary idea from a problem context and one part of the idea. Models trained on \kbname substantially improved in this task when evaluated on held-out recombinations. We hope \kbname provides a foundation for future work on cross-disciplinary ideation systems, and unlocks new empirical studies on innovation at scale.

\section*{Limitations}\label{limitstions}
\paragraph{Extraction quality} \revision{As with any automatically-extracted knowledge base, \kbname naturally contains some extraction errors (see Appendix \ref{sec:qualitative_analysis}). Importantly, our work is the first to explore the new task of extracting recombinations from papers, revealing a gap between the performance of extraction models and humans on the task. As is the case with newly-introduced NLP tasks, future methods trained on our annotated corpus are expected to further improve extraction results, and hence the quality of \kbname. However, our analysis shows we already reach good extraction quality overall with minor errors (see Section \ref{par:extraction_accuracy}), and our downstream applications further demonstrate that the data in \kbname can be used to derive utility in scientific meta-analysis and ideation.}

\paragraph{Abstract-level scope} \revision{\kbname focuses on extracting recombination instances from scientific abstracts rather than full papers. This design choice, common in scientific IE tasks \cite{gonzalez2023beyond, zhang2024massw, naik2024care}, enables more scalable annotation and leverages the fact that abstracts typically summarize key contributions—including conceptual recombinations. In our setting, we focus on capturing cases where a recombination is at the core of a paper's contribution, hence likely to appear in the abstract. Confirming this intuition, we further conduct an analysis that finds that abstracts cover the vast majority of these cases.  Extending extraction methods to full papers could reveal additional recombination patterns in future work.}

\paragraph{Recombination prediction evaluation}
\revision{As in other open-ended creative tasks \cite{Jentzsch2023ChatGPTIFA, Meng2023FollowupQGTI, Huot2024AgentsRN}, the recombination prediction task admits no single correct answer. Given a problem description, there are many valid ways to blend ideas or draw inspiration, which can lead to false negatives and an overly conservative estimate of model performance. To mitigate this, we conduct a complementary human evaluation. However, due to the expertise required from evaluators, the scale and depth of this assessment are necessarily limited.}

\paragraph{Experimenting with additional models} Our work leverages a diverse set of models for extraction and prediction, including open-source LLMs (e.g., \texttt{Mistral-7B}, \texttt{all-mpnet-base}), proprietary models (e.g., \texttt{GPT-4o}), and non-generative baselines (e.g., \texttt{PURE}). \texttt{GPT-4o} is used for auxiliary tasks, such as evaluation (judging entity span similarity), analysis (identifying entity's scientific domain), and to enrich our data (generating a context string for the extracted recombinations). As our primary focus is on building and analyzing the recombination knowledge base, we limit our experiments to these models. Exploring a broader range of models for these auxiliary tasks is an important direction for future work.

\section*{Ethical Considerations}
To collect human-annotated recombination examples, we recruited crowdworkers through the Upwork platform. All annotators were informed in advance about the nature, purpose, and scope of the annotation task. They were compensated fairly for their time, at rates ranging from \$26 to \$30 per hour. Annotation quality was monitored through overlapping assignments and expert review to ensure reliability and accuracy.

For our human evaluation study, three volunteers with prior research experience participated in ranking model outputs. Participation was entirely voluntary, and no personal or identifying information about the annotators or participants is collected or disclosed.

To support transparency and reproducibility, we release our code, model checkpoints, and the annotated data under an open license. We used AI-based coding assistants (e.g., GitHub Copilot) and language tools for minor code and grammar refinements during development.

\bibliography{custom}

@article{Saier2020unarXiveAL,
  title={unarXive: a large scholarly data set with publications’ full-text, annotated in-text citations, and links to metadata},
  author={Tarek Saier and Michael F{\"a}rber},
  journal={Scientometrics},
  year={2020},
  volume={125},
  pages={3085 - 3108},
  url={https://api.semanticscholar.org/CorpusID:211574327}
}

@inproceedings{choi2024creativeconnect,
  title={Creativeconnect: Supporting reference recombination for graphic design ideation with generative ai},
  author={Choi, DaEun and Hong, Sumin and Park, Jeongeon and Chung, John Joon Young and Kim, Juho},
  booktitle={Proceedings of the 2024 CHI Conference on Human Factors in Computing Systems},
  pages={1--25},
  year={2024}
}

@article{falkenhainer1989structure,
  title={The structure-mapping engine: Algorithm and examples},
  author={Falkenhainer, Brian and Forbus, Kenneth D and Gentner, Dedre},
  journal={Artificial intelligence},
  volume={41},
  number={1},
  pages={1--63},
  year={1989},
  publisher={Elsevier}
}

@inproceedings{Perry2021LightTagTA,
  title={LightTag: Text Annotation Platform},
  author={Tal Perry},
  booktitle={Conference on Empirical Methods in Natural Language Processing},
  year={2021},
  url={https://api.semanticscholar.org/CorpusID:237420609}
}

@inproceedings{Zhong2021AFE,
  title={A Frustratingly Easy Approach for Entity and Relation Extraction},
  author={Zexuan Zhong and Danqi Chen},
  booktitle={North American Chapter of the Association for Computational Linguistics},
  year={2021},
  url={https://api.semanticscholar.org/CorpusID:261317734}
}

@article{Hu2021LoRALA,
  title={LoRA: Low-Rank Adaptation of Large Language Models},
  author={J. Edward Hu and Yelong Shen and Phillip Wallis and Zeyuan Allen-Zhu and Yuanzhi Li and Shean Wang and Weizhu Chen},
  journal={ArXiv},
  year={2021},
  volume={abs/2106.09685},
  url={https://api.semanticscholar.org/CorpusID:235458009}
}

@inproceedings{Sharif2024ExplicitIA,
  title={Explicit, Implicit, and Scattered: Revisiting Event Extraction to Capture Complex Arguments},
  author={Omar Sharif and Joseph Gatto and Madhusudan Basak and Sarah Masud Preum},
  booktitle={Conference on Empirical Methods in Natural Language Processing},
  year={2024},
  url={https://api.semanticscholar.org/CorpusID:273162986}
}

@article{Shi2019SurprisingCO,
  title={Surprising combinations of research contents and contexts are related to impact and emerge with scientific outsiders from distant disciplines},
  author={Feng Shi and James Allen Evans},
  journal={Nature Communications},
  year={2019},
  volume={14},
  url={https://api.semanticscholar.org/CorpusID:204800519}
}

@article{Chan2018SOLVENT,
  title={SOLVENT},
  author={Joel Chan and Joseph Chee Chang and Tom Hope and Dafna Shahaf and Aniket Kittur},
  journal={Proceedings of the ACM on Human-Computer Interaction},
  year={2018},
  volume={2},
  pages={1 - 21},
  url={https://api.semanticscholar.org/CorpusID:53236685}
}

@article{Krenn2022ForecastingTF,
  title={Forecasting the future of artificial intelligence with machine learning-based link prediction in an exponentially growing knowledge network},
  author={Mario Krenn and Lorenzo Buffoni and Bruno Coutinho and Sagi Eppel and Jacob Gates Foster and Andrew Gritsevskiy and Harlin Lee and Yichao Lu and Jo{\~a}o P. Moutinho and Nima Sanjabi and Rishi Sonthalia and Ngoc M. Tran and Francisco Valente and Yangxinyu Xie and Rose Yu and Michael Kopp},
  journal={Nature Machine Intelligence},
  year={2022},
  volume={5},
  pages={1326-1335},
  url={https://api.semanticscholar.org/CorpusID:252682946}
}

@article{Choi2023CreativeConnectSR,
  title={CreativeConnect: Supporting Reference Recombination for Graphic Design Ideation with Generative AI},
  author={DaEun Choi and Sumin Hong and Jeongeon Park and John Joon Young Chung and Juho Kim},
  journal={Proceedings of the CHI Conference on Human Factors in Computing Systems},
  year={2023},
  url={https://api.semanticscholar.org/CorpusID:266362188}
}

@article{Suh2023LuminateSG,
  title={Luminate: Structured Generation and Exploration of Design Space with Large Language Models for Human-AI Co-Creation},
  author={Sangho Suh and Meng Chen and Bryan Min and Toby Jia-Jun Li and Haijun Xia},
  journal={Proceedings of the CHI Conference on Human Factors in Computing Systems},
  year={2023},
  url={https://api.semanticscholar.org/CorpusID:269752103}
}

@inproceedings{Holyoak1994MentalLA,
  title={Mental Leaps: Analogy in Creative Thought},
  author={Keith J. Holyoak and Paul Thagard},
  year={1994},
  url={https://api.semanticscholar.org/CorpusID:143622641}
}

@inproceedings{2019TheCH,
  title={The Cambridge Handbook of Creativity},
  author={},
  year={2019},
  url={https://api.semanticscholar.org/CorpusID:164199851}
}

@article{McKeown2014TheCS,
  title={The cognitive science of science: explanation, discovery, and conceptual change},
  author={C{\'e}line McKeown},
  journal={Ergonomics},
  year={2014},
  volume={57},
  pages={632 - 633},
  url={https://api.semanticscholar.org/CorpusID:109591540}
}

@article{Chilton2019VisiBlendsAF,
  title={VisiBlends: A Flexible Workflow for Visual Blends},
  author={Lydia B. Chilton and S. Petridis and Maneesh Agrawala},
  journal={Proceedings of the 2019 CHI Conference on Human Factors in Computing Systems},
  year={2019},
  url={https://api.semanticscholar.org/CorpusID:96456140}
}

@article{Srinivasan2024ImprovingSO,
  title={Improving Selection of Analogical Inspirations through Chunking and Recombination},
  author={Arvind Srinivasan and Joel Chan},
  journal={Proceedings of the 16th Conference on Creativity \& Cognition},
  year={2024},
  url={https://api.semanticscholar.org/CorpusID:270638960}
}

@article{Luan2018MultiTaskIO,
  title={Multi-Task Identification of Entities, Relations, and Coreference for Scientific Knowledge Graph Construction},
  author={Yi Luan and Luheng He and Mari Ostendorf and Hannaneh Hajishirzi},
  journal={ArXiv},
  year={2018},
  volume={abs/1808.09602},
  url={https://api.semanticscholar.org/CorpusID:52118895}
}

@inproceedings{Jain2020SciREXAC,
  title={SciREX: A Challenge Dataset for Document-Level Information Extraction},
  author={Sarthak Jain and Madeleine van Zuylen and Hannaneh Hajishirzi and Iz Beltagy},
  booktitle={Annual Meeting of the Association for Computational Linguistics},
  year={2020},
  url={https://api.semanticscholar.org/CorpusID:218470122}
}

@article{Sainz2023GoLLIEAG,
  title={GoLLIE: Annotation Guidelines improve Zero-Shot Information-Extraction},
  author={Oscar Sainz and Iker Garc{\'i}a-Ferrero and Rodrigo Agerri and Oier L{\'o}pez de Lacalle and German Rigau and Eneko Agirre},
  journal={ArXiv},
  year={2023},
  volume={abs/2310.03668},
  url={https://api.semanticscholar.org/CorpusID:263671572}
}

@inproceedings{Hennen2024ITERIT,
  title={ITER: Iterative Transformer-based Entity Recognition and Relation Extraction},
  author={Moritz Hennen and Florian Babl and Michaela Geierhos},
  booktitle={Conference on Empirical Methods in Natural Language Processing},
  year={2024},
  url={https://api.semanticscholar.org/CorpusID:274060213}
}

@misc{bge_embedding,
      title={C-Pack: Packaged Resources To Advance General Chinese Embedding}, 
      author={Shitao Xiao and Zheng Liu and Peitian Zhang and Niklas Muennighoff},
      year={2023},
      eprint={2309.07597},
      archivePrefix={arXiv},
      primaryClass={cs.CL}
}

@inproceedings{Teach2020YouCT,
  title={You CAN Teach an Old Dog New Tricks! On Training Knowledge Graph Embeddings},
  author={You Can Teach and Daniel Ruffinelli and Samuel Broscheit and Rainer Gemulla},
  booktitle={International Conference on Learning Representations},
  year={2020},
  url={https://api.semanticscholar.org/CorpusID:211241737}
}

@article{Sun2023IsCG,
  title={Is ChatGPT Good at Search? Investigating Large Language Models as Re-Ranking Agent},
  author={Weiwei Sun and Lingyong Yan and Xinyu Ma and Pengjie Ren and Dawei Yin and Zhaochun Ren},
  journal={ArXiv},
  year={2023},
  volume={abs/2304.09542},
  url={https://api.semanticscholar.org/CorpusID:258212638}
}

@article{wei2022chain,
  title={Chain-of-thought prompting elicits reasoning in large language models},
  author={Wei, Jason and Wang, Xuezhi and Schuurmans, Dale and Bosma, Maarten and Xia, Fei and Chi, Ed and Le, Quoc V and Zhou, Denny and others},
  journal={Advances in Neural Information Processing Systems},
  volume={35},
  pages={24824--24837},
  year={2022}
}

@article{kang2022augmenting,
  title={Augmenting scientific creativity with an analogical search engine},
  author={Kang, Hyeonsu B and Qian, Xin and Hope, Tom and Shahaf, Dafna and Chan, Joel and Kittur, Aniket},
  journal={ACM Transactions on Computer-Human Interaction},
  volume={29},
  number={6},
  pages={1--36},
  year={2022},
  publisher={ACM New York, NY}
}

@article{wang2022text,
  title={Text embeddings by weakly-supervised contrastive pre-training},
  author={Wang, Liang and Yang, Nan and Huang, Xiaolong and Jiao, Binxing and Yang, Linjun and Jiang, Daxin and Majumder, Rangan and Wei, Furu},
  journal={arXiv preprint arXiv:2212.03533},
  year={2022}
}

@String{Computing = "Computing" }

@String{Academic = "Academic Press" }

@inproceedings{naik2024care,
  title={CARE: Extracting Experimental Findings From Clinical Literature},
  author={Naik, Aakanksha and Kuehl, Bailey and Bransom, Erin and Downey, Doug and Hope, Tom},
  booktitle={Findings of the Association for Computational Linguistics: NAACL 2024},
  pages={4580--4596},
  year={2024}
}

@article{radensky2024scideator,
  title={Scideator: Human-LLM Scientific Idea Generation Grounded in Research-Paper Facet Recombination},
  author={Radensky, Marissa and Shahid, Simra and Fok, Raymond and Siangliulue, Pao and Weld, Daniel S and Hope, Tom},
  journal={arXiv preprint arXiv:2409.14634},
  year={2024}
}

@article{scimon,
  title={SciMON: Scientific Inspiration Machines Optimized for Novelty},
  author={Wang, Qingyun and Downey, Doug and Ji, Heng and Hope, Tom},
  journal={ACL},
  year={2024}
}

@inproceedings{hope2017KDD,
 author = {Hope, Tom and Chan, Joel and Kittur, Aniket and Shahaf, Dafna},
 title = {Accelerating Innovation Through Analogy Mining},
 booktitle = {Proceedings of the 23rd ACM SIGKDD International Conference on Knowledge Discovery and Data Mining},
 series = {KDD '17},
 year = {2017},
 isbn = {978-1-4503-4887-4},
 location = {Halifax, NS, Canada},
 pages = {235--243},
 numpages = {9},
 url = {http://doi.acm.org/10.1145/3097983.3098038},
 doi = {10.1145/3097983.3098038},
 acmid = {3098038},
 publisher = {ACM},
 address = {New York, NY, USA},
}

@article{shi2023surprising,
  title={Surprising combinations of research contents and contexts are related to impact and emerge with scientific outsiders from distant disciplines},
  author={Shi, Feng and Evans, James},
  journal={Nature Communications},
  volume={14},
  number={1},
  pages={1641},
  year={2023},
  publisher={Nature Publishing Group UK London}
}

@article{gentner1997analogy,
  title={Analogy and creativity in the works of Johannes Kepler},
  author={Gentner, Dedre and Brem, Sarah and Ferguson, Ron and Wolff, Philip and Markman, Arthur B and Forbus, KD},
  journal={Creative thought: An investigation of conceptual structures and processes},
  pages={403--459},
  year={1997},
  publisher={Citeseer}
}

@article{chan2011benefits,
  title={On the benefits and pitfalls of analogies for innovative design: Ideation performance based on analogical distance, commonness, and modality of examples},
  author={Chan, Joel and Fu, Katherine and Schunn, Christian and Cagan, Jonathan and Wood, Kristin and Kotovsky, Kenneth},
  journal={Journal of mechanical design},
  volume={133},
  number={8},
  pages={081004},
  year={2011},
  publisher={American Society of Mechanical Engineers}
}

@article{gentner1983structure,
  title={Structure-Mapping: A Theoretical Framework for Analogy},
  author={Gentner, Dedre},
  journal={Cognitive science},
  year={1983},
}

@article{knoblichConstraintRelaxationChunk1999,
	title = {Constraint relaxation and chunk decomposition in insight problem solving},
	volume = {25},
	number = {6},
	journal = {Journal of Experimental Psychology: Learning, Memory, and Cognition},
	author = {Knoblich, G. and Ohlsson, S. and Haider, H. and Rhenius, D.},
	year = {1999},
	note = {00691},
	pages = {1534--1555},
}

@article{mccaffreyInnovationReliesObscure2012,
	title = {Innovation {Relies} on the {Obscure}: {A} {Key} to {Overcoming} the {Classic} {Problem} of {Functional} {Fixedness}},
	volume = {23},
	issn = {1467-9280},
	number = {3},
	journal = {Psychological Science},
	author = {McCaffrey, T.},
	year = {2012},
	note = {00117},
	pages = {215--218},
}

@article{uzzi2013atypical,
  title={Atypical combinations and scientific impact},
  author={Uzzi, Brian and Mukherjee, Satyam and Stringer, Michael and Jones, Ben},
  journal={Science},
  volume={342},
  number={6157},
  pages={468--472},
  year={2013},
  publisher={American Association for the Advancement of Science}
}

@article{sosa2022contexts,
  title={Contexts and contradictions: a roadmap for computational drug repurposing with knowledge inference},
  author={Sosa, Daniel N and Altman, Russ B},
  journal={Briefings in Bioinformatics},
  volume={23},
  number={4},
  pages={bbac268},
  year={2022},
  publisher={Oxford University Press},
  url={https://pubmed.ncbi.nlm.nih.gov/35817308/}
}

@inproceedings{Pan2024SciDMTAL,
  title={SciDMT: A Large-Scale Corpus for Detecting Scientific Mentions},
  author={Huitong Pan and Qi Zhang and Cornelia Caragea and Eduard Constantin Dragut and Longin Jan Latecki},
  booktitle={International Conference on Language Resources and Evaluation},
  year={2024},
  url={https://api.semanticscholar.org/CorpusID:269804233}
}

@article{yang2025moose,
  title={MOOSE-Chem2: Exploring LLM Limits in Fine-Grained Scientific Hypothesis Discovery via Hierarchical Search},
  author={Yang, Zonglin and Liu, Wanhao and Gao, Ben and Liu, Yujie and Li, Wei and Xie, Tong and Bing, Lidong and Ouyang, Wanli and Cambria, Erik and Zhou, Dongzhan},
  journal={arXiv preprint arXiv:2505.19209},
  year={2025}
}

@misc{yang2025moosechemlargelanguagemodels,
      title={MOOSE-Chem: Large Language Models for Rediscovering Unseen Chemistry Scientific Hypotheses}, 
      author={Zonglin Yang and Wanhao Liu and Ben Gao and Tong Xie and Yuqiang Li and Wanli Ouyang and Soujanya Poria and Erik Cambria and Dongzhan Zhou},
      year={2025},
      eprint={2410.07076},
      archivePrefix={arXiv},
      primaryClass={cs.CL},
      url={https://arxiv.org/abs/2410.07076}, 
}

@article{Huot2024AgentsRN,
  title={Agents' Room: Narrative Generation through Multi-step Collaboration},
  author={Fantine Huot and Reinald Kim Amplayo and Jennimaria Palomaki and Alice Shoshana Jakobovits and Elizabeth Clark and Mirella Lapata},
  journal={ArXiv},
  year={2024},
  volume={abs/2410.02603},
  url={https://api.semanticscholar.org/CorpusID:273098211}
}

@inproceedings{Meng2023FollowupQGTI,
  title={FollowupQG: Towards information-seeking follow-up question generation},
  author={Yan Meng and Liangming Pan and Yixin Cao and Min-Yen Kan},
  booktitle={International Joint Conference on Natural Language Processing},
  year={2023},
  url={https://api.semanticscholar.org/CorpusID:261681892}
}

@article{Jentzsch2023ChatGPTIFA,
  title={ChatGPT is fun, but it is not funny! Humor is still challenging Large Language Models},
  author={Sophie F. Jentzsch and K. Kersting},
  journal={ArXiv},
  year={2023},
  volume={abs/2306.04563},
  url={https://api.semanticscholar.org/CorpusId:259095915}
}

@article{zhang2024massw,
  title={Massw: A new dataset and benchmark tasks for ai-assisted scientific workflows},
  author={Zhang, Xingjian and Xie, Yutong and Huang, Jin and Ma, Jinge and Pan, Zhaoying and Liu, Qijia and Xiong, Ziyang and Ergen, Tolga and Shim, Dongsub and Lee, Honglak and others},
  journal={arXiv preprint arXiv:2406.06357},
  year={2024}
}

@article{gonzalez2023beyond,
  title={Beyond good intentions: Reporting the research landscape of nlp for social good},
  author={Gonzalez, Fernando and Jin, Zhijing and Sch{\"o}lkopf, Bernhard and Hope, Tom and Sachan, Mrinmaya and Mihalcea, Rada},
  journal={arXiv preprint arXiv:2305.05471},
  year={2023}
}

@article{Frohnert2024DiscoveringECA,
  title={Discovering emergent connections in quantum physics research via dynamic word embeddings},
  author={Felix Frohnert and Xuemei Gu and Mario Krenn and Evert P L van Nieuwenburg},
  journal={Machine Learning: Science and Technology},
  year={2024},
  volume={6},
  url={https://api.semanticscholar.org/CorpusId:273963065}
}

@article{delgado2019cohen,
  title={Why Cohen’s Kappa should be avoided as performance measure in classification},
  author={Delgado, Rosario and Tibau, Xavier-Andoni},
  journal={PloS one},
  volume={14},
  number={9},
  pages={e0222916},
  year={2019},
  publisher={Public Library of Science San Francisco, CA USA}
}

@inproceedings{pramanick-etal-2025-nature,
    title = "The Nature of {NLP}: Analyzing Contributions in {NLP} Papers",
    author = "Pramanick, Aniket  and
      Hou, Yufang  and
      Mohammad, Saif M.  and
      Gurevych, Iryna",
    editor = "Che, Wanxiang  and
      Nabende, Joyce  and
      Shutova, Ekaterina  and
      Pilehvar, Mohammad Taher",
    booktitle = "Proceedings of the 63rd Annual Meeting of the Association for Computational Linguistics (Volume 1: Long Papers)",
    month = jul,
    year = "2025",
    address = "Vienna, Austria",
    publisher = "Association for Computational Linguistics",
    url = "https://aclanthology.org/2025.acl-long.1224/",
    pages = "25169--25191",
    ISBN = "979-8-89176-251-0"
}

@article{Wahle2023WeAW,
  title={We are Who We Cite: Bridges of Influence Between Natural Language Processing and Other Academic Fields},
  author={Jan Philip Wahle and Terry Ruas and Mohamed Abdalla and Bela Gipp and Saif Mohammad},
  journal={ArXiv},
  year={2023},
  volume={abs/2310.14870},
  url={https://api.semanticscholar.org/CorpusID:264436427}
}

@article{si2024can,
  title={Can llms generate novel research ideas? a large-scale human study with 100+ nlp researchers},
  author={Si, Chenglei and Yang, Diyi and Hashimoto, Tatsunori},
  journal={arXiv preprint arXiv:2409.04109},
  year={2024}
}

@misc{garikaparthi2025irisinteractiveresearchideation,
      title={IRIS: Interactive Research Ideation System for Accelerating Scientific Discovery}, 
      author={Aniketh Garikaparthi and Manasi Patwardhan and Lovekesh Vig and Arman Cohan},
      year={2025},
      eprint={2504.16728},
      archivePrefix={arXiv},
      primaryClass={cs.AI},
      url={https://arxiv.org/abs/2504.16728}, 
}

@article{Fortunato2018ScienceOS,
  title={Science of Science},
  author={Santo Fortunato and Carl T. Bergstrom and Katy B{\"o}rner and James A. Evans and Dirk Helbing and Stasa Milojevic and Alexander Michael Petersen and Filippo Radicchi and Roberta Sinatra and Brian Uzzi and Alessandro Vespignani and Ludo Waltman and Dashun Wang and Albert-Ĺaszl{\'o} Barab{\'a}si},
  journal={Nature},
  year={2018},
  volume={214},
  pages={1-2},
  url={https://api.semanticscholar.org/CorpusID:3637715}
}

@misc{katz2024knowledgenavigatorllmguidedbrowsing,
      title={Knowledge Navigator: LLM-guided Browsing Framework for Exploratory Search in Scientific Literature}, 
      author={Uri Katz and Mosh Levy and Yoav Goldberg},
      year={2024},
      eprint={2408.15836},
      archivePrefix={arXiv},
      primaryClass={cs.IR},
      url={https://arxiv.org/abs/2408.15836}, 
}

\appendix

\section*{Appendix Contents}
\vspace{-0.5em}
\noindent\textbf{\ref{sec:IAA}~~Annotator Agreement} \dotfill \pageref{sec:IAA}\\
\hspace*{1.5em}\ref{subsec:disagreement}~~Disagreement Analysis \dotfill \pageref{subsec:disagreement}\\
\noindent\textbf{\ref{sec:extraction_baselines_parent}~~Additional Extraction Details} \dotfill \pageref{sec:extraction_baselines_parent}\\
\hspace*{1.5em}\ref{sec:recomb_keywords}~~Recombination Keywords \dotfill \pageref{sec:recomb_keywords}\\
\hspace*{1.5em}\ref{sec:extraction_baselines}~~Extraction Baselines Implementation \dotfill \pageref{sec:extraction_baselines}\\
\hspace*{1.5em}\ref{sec:e2e_vs_spacial}~~E2E vs.\ Specialized Extraction \dotfill \pageref{sec:e2e_vs_spacial}\\
\hspace*{1.5em}\ref{sec:span_sim_prompt}~~Span Similarity \dotfill \pageref{sec:span_sim_prompt}\\
\hspace*{1.5em}\ref{sec:qualitative_analysis}~~Extraction Error Analysis \dotfill \pageref{sec:qualitative_analysis}\\
\hspace*{1.5em}\ref{section:qualitative_extraction_eval}~~Large-Scale Extraction Assessment \dotfill \pageref{section:qualitative_extraction_eval}\\
\noindent\textbf{\ref{sec:kg_const}~~Additional KG Construction Details} \dotfill \pageref{sec:kg_const}\\
\hspace*{1.5em}\ref{sec:domain_anlysis}~~Graph Node Domains \dotfill \pageref{sec:domain_anlysis}\\
\hspace*{1.5em}\ref{sec:entity_norm}~~Entity Normalization \dotfill \pageref{sec:entity_norm}\\
\hspace*{1.5em}\ref{sec:postprocessing}~~Entity Postprocessing \dotfill \pageref{sec:postprocessing}\\
\noindent\textbf{\ref{sec:graph_analysis}~~Additional Knowledge Base Analysis} \dotfill \pageref{sec:graph_analysis}\\
\hspace*{1.5em}\ref{sec:keyword_freq}~~Keywords in Annotated Data \dotfill \pageref{sec:keyword_freq}\\
\hspace*{1.5em}\ref{sec:predominant_recomb}~~Predominant Recombination Relations \dotfill \pageref{sec:predominant_recomb}\\
\hspace*{1.5em}\ref{sec:fine_grained_types}~~Nuanced Recombination Types \dotfill \pageref{sec:fine_grained_types}\\
\noindent\textbf{\ref{sec:pred_details}~~Additional Prediction Details} \dotfill \pageref{sec:pred_details}\\
\hspace*{1.5em}\ref{sec:pred_data_prep}~~Prediction Data Preprocessing \dotfill \pageref{sec:pred_data_prep}\\
\hspace*{1.5em}\ref{sec:prediction_baselines}~~Prediction Baselines \dotfill \pageref{sec:prediction_baselines}\\
\hspace*{1.5em}\ref{sec:reranker_errors}~~Reranker Error Analysis \dotfill \pageref{sec:reranker_errors}\\
\hspace*{1.5em}\ref{sec:pred_examples}~~Prediction Examples \dotfill \pageref{sec:pred_examples}\\
\noindent\textbf{\ref{app:user_study}~~User Study Additional Details} \dotfill \pageref{app:user_study}\\
\noindent\textbf{\ref{sec:extraction_comp}~~Comparison to Other IE Methods} \dotfill \pageref{sec:extraction_comp}\\

\vspace{0.8em}
\noindent\textbf{List of Prompts}\\[0.3em]
\hspace*{1.5em}Figure~\ref{fig:Extraction-Prompt}~~E2E Extraction \dotfill \pageref{fig:Extraction-Prompt}\\
\hspace*{1.5em}Figure~\ref{fig:e2e_icl_prompt}~~E2E ICL  \dotfill \pageref{fig:e2e_icl_prompt}\\
\hspace*{1.5em}Figure~\ref{fig:ner_icl_prompt}~~Entity Extraction \dotfill \pageref{fig:ner_icl_prompt}\\
\hspace*{1.5em}Figure~\ref{fig:span_similarity_prompt}~~Span Similarity  \dotfill \pageref{fig:span_similarity_prompt}\\
\hspace*{1.5em}Figure~\ref{fig:large_scale_prompt}~~Large-Scale Evaluation \dotfill \pageref{fig:large_scale_prompt}\\
\hspace*{1.5em}Figure~\ref{fig:comb_domain_prompt}~~Blend Domain Analysis \dotfill \pageref{fig:comb_domain_prompt}\\
\hspace*{1.5em}Figure~\ref{fig:inspo_domain_prompt}~~Inspiration Domain Analysis \dotfill \pageref{fig:inspo_domain_prompt}\\
\hspace*{1.5em}Figure~\ref{fig:context_extraction_prompt}~~Context Extraction \dotfill \pageref{fig:context_extraction_prompt}\\
\hspace*{1.5em}Figure~\ref{fig:leakage_detect_prompt}~~Leak Detection \dotfill \pageref{fig:leakage_detect_prompt}\\
\hspace*{1.5em}Figure~\ref{fig:rankgpt_prompt}~~Adjusted RankGPT \dotfill \pageref{fig:rankgpt_prompt}\\
\hspace*{1.5em}Figure~\ref{fig:postprocessing_prompt}~~Entity Postprocessing  \dotfill \pageref{fig:postprocessing_prompt}\\

\section{Annotator Agreement}\label{sec:IAA}

\revision{Following standard practice in information extraction \cite{Naik2023CAREEE, Sharif2024ExplicitIA}, we assess inter-annotator agreement using precision, recall, and F1 scores. Agreement is computed by treating one annotator’s labels as the reference and the other’s as predictions. In addition to measuring entity-level and relation-level agreement, we also evaluate agreement on recombination presence—that is, whether a text expresses a recombination instance, regardless of its type.}

\revision{We apply the soft entity and relation matching procedure described in Section~\ref{eval_criteria} to compute entity and relation agreement. All agreement scores are based on the 49 documents annotated by both annotators (approximately 10\% of the full dataset). We treat these agreement measures as a proxy for human-level performance on this task.}

\subsection{Disagreement Analysis}\label{subsec:disagreement}
\begin{table*}[!htbp]
    \centering
    \footnotesize
    \begin{tabular}{@{}p{15cm}@{}}
    \toprule
    \multicolumn{1}{c}{\textbf{Annotators' Disagreement Examples}}\\
    \midrule
    \textbf{Abstract}: "…This research proposed a framework based on Long Short-Term Memory (LSTM) deep learning network to generate day-ahead hourly temperature forecast… A case study is shown which uses historical in-situ observations and \marker{Internet of Things (IoT) observations for New York City, USA}. By leveraging the \marker{historical air temperature data from in-situ observations}, the LSTM model can be exposed to more historical patterns that might not be present in the IoT observations. Meanwhile, by using IoT observations, the spatial resolution of air temperature predictions is significantly improved..."\newline\newline
    \textbf{Annotator 1}: [\textbf{Blend}: "\textit{Internet of Things (IoT) observations for New York City, USA}" $\longleftrightarrow$ "\textit{historical air temperature data from in-situ observations}"] \newline
    \textbf{Annotator 2}:  []  \newline\newline
    \textbf{Resolution}: Upon expert review, Annotator 1’s interpretation was selected, as the authors explicitly describe how the two sources of data serve complementary roles in their method.\\ \midrule
        \textbf{Abstract}: "...we propose an integrated system that can perform large-scale autonomous flights and real-time semantic mapping in challenging under-canopy environments. We detect and model tree trunks and ground planes from \marker{LiDAR data}, which are associated across scans and used to constrain robot poses as well as tree trunk models. The autonomous navigation module utilizes \marker{a multi-level planning and mapping framework} and computes dynamically feasible trajectories that lead the UAV to build a semantic map of the user-defined region of interest in a computationally and storage efficient manner. A drift-compensation mechanism is designed to minimize the odometry drift using semantic SLAM outputs in real time, while maintaining planner optimality and controller stability..."\newline\newline
    \textbf{Annotator 1}: [\textbf{Blend}: "\textit{LiDAR data}" $\longleftrightarrow$ "\textit{a multi-level planning and mapping framework}"] \newline
    \textbf{Annotator 2}:  []  \newline\newline
    \textbf{Resolution}: Annotator 2’s judgment was selected after expert review, as the relation between the two components is not clearly described as a recombination.\\
    \bottomrule
    \end{tabular}
    \caption{\revision{Examples of annotation disagreements and resolutions by expert review.}}\label{tab:disagreement_examples}
\end{table*}

\revision{To better understand the sources of annotation disagreement, we conducted a qualitative analysis. The most common cause stems from differences in identifying whether a recombination is present at all (see examples in Table~\ref{tab:disagreement_examples}). In such cases, disagreements were resolved via discussion and expert adjudication. The primary criterion for resolution was whether the authors explicitly describe a recombination as contributing to their approach.}

\revision{Interestingly, once annotators agreed that a recombination was present, they rarely disagreed on its type, and only a single example exhibited this form of conflict. However, disagreements over which entities the recombination includes were more frequent. These typically fell into two categories:} 
 \begin{enumerate}
     \item \revision{\textbf{Boundary disagreements}, where annotators selected different spans with overlapping meaning. Here, the expert favored the span that preserved more context (e.g., "\textit{reinforcement learning which uses traditional time series stock price data}" was preferred over "\textit{traditional time series stock price data}").} 
     \item \revision{\textbf{Conceptual disagreements}, where annotators identified fundamentally different entities. These were resolved through further discussion and clarification.}
 \end{enumerate}

\section{Additional Extraction Details}\label{sec:extraction_baselines_parent}
\subsection{Recombination keywords}\label{sec:recomb_keywords}
We use keyword-based filtering to identify works that are more likely to discuss recombination before assigning papers to human annotators. Table \ref{table:keywords} presents the list of keywords used for this step.

\begin{table*}[!hbtp]
\centering
\footnotesize
\begin{tabular}{@{}llllll@{}}
\toprule
\multicolumn{6}{c}{\textbf{Recombination keywords}}\\
\midrule
combines	&	analogies	&	aggregate	&	intermingle	&	unify	&	blending	\\
combined	&	equivalence	&	aggregation	&	intermingling	&	unification	&	blends	\\
combine	&	equivalent	&	align	&	join	&	weave	&	blend	\\
combination	&	reduction	&	alignment	&	joining	&	weaving	&	blends	\\
combinations	&	reframing	&	amalgamate	&	juxtapose	&	hybrid	&	merge	\\
combining	&	reframe	&	amalgamation	&	juxtaposition	&	merge	&	merges	\\
mixing	&	reformulating	&	assemble	&	link	&	merges	&	unites	\\
mixture	&	casting	&	assembling	&	linkage	&	merging	&	analogy	\\
mix	&	cast	&	associate	&	meld	&	merged	&	analogize	\\
mixed	&	casts	&	association	&	melding	&	conflation	&	analogies	\\
integrates	&	viewing	&	bond	&	mesh	&	couple	&	equivalence	\\
integrating	&	viewed	&	bonding	&	meshing	&	unite	&	equivalent	\\
integrate	&	view	&	bridge	&	perceive	&	unites	&	correlate	\\
integrated	&	inspire	&	bridging	&	perception	&	interplay	&	correlation	\\
connection	&	inspired	&	coalesce	&	relate	&	interconnect	&	envision	\\
synergy	&	inspiration	&	coalescence	&	relation	&	harmonize	&	envisioning	\\
fusion	&	inspires	&	compose	&	splice	&	harmony	&	harmonize	\\
fuses	&	inspiring	&	composition	&	splicing	&	incorporate	&	harmony	\\
unify	&	interconnect	&	incorporation	&	synthesis	&	reduction	&	synthesis	\\
aggregate	&	align	&	inspiring	&	inspire	&	couple	&	conjunction	\\
aggregation	&	reframing	&	inspiration	&	fuse	&	unite	&	conjoin	\\
alignment	&	reframe	&	inspires	&	synthesis	&		&		\\
\bottomrule
\end{tabular}
\caption{Recombination keywords. We use a predefined list of keywords to identify works that are more likely to discuss idea recombination.}\label{table:keywords}%
\end{table*}

\subsection{Extraction baselines implementation}\label{sec:extraction_baselines}
\begin{figure*}[!htbp]
    \centering
    \includegraphics[width=\textwidth]{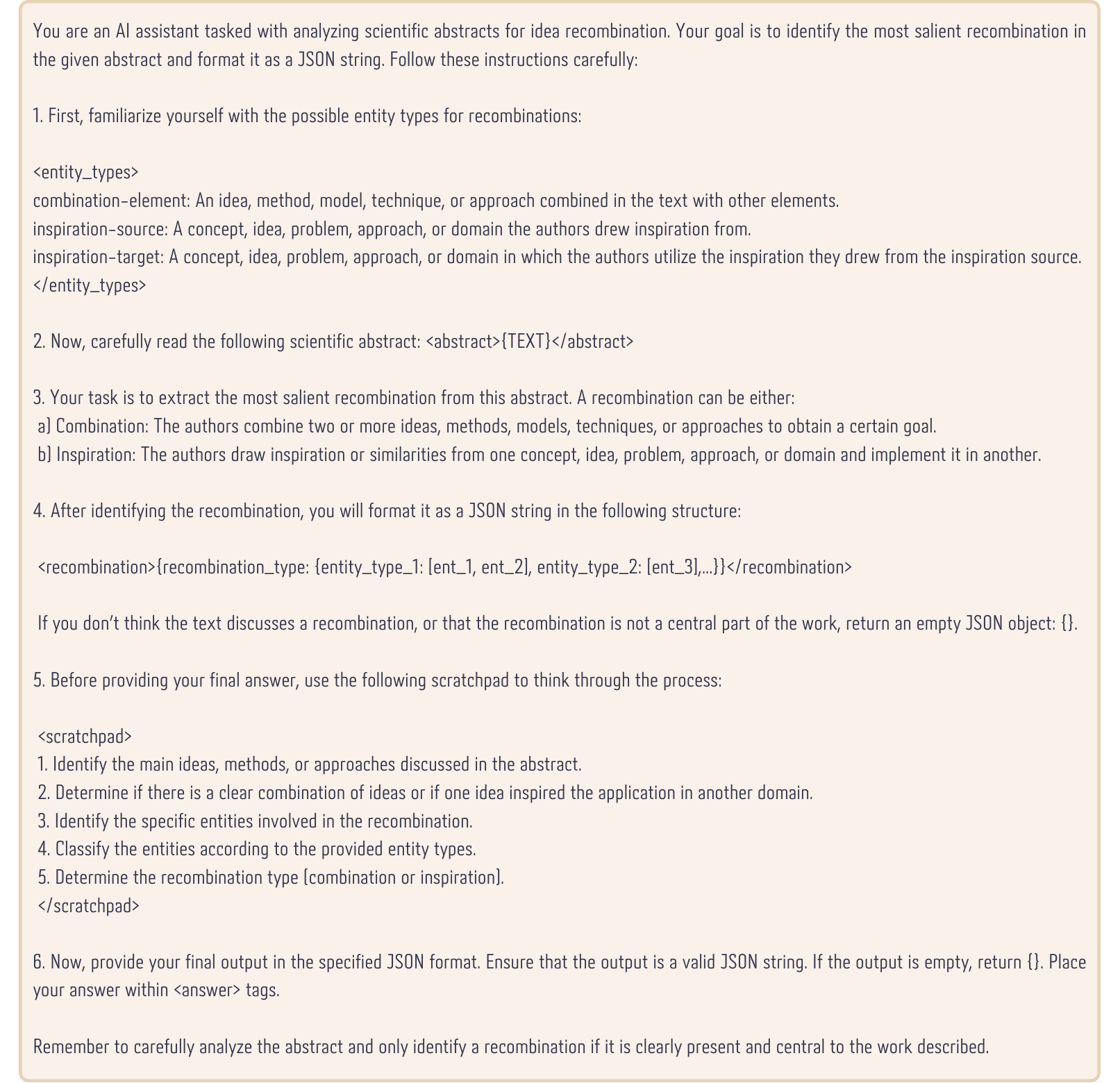}
    \caption{E2E extraction prompt. \{TEXT\} is the placeholder for the input abstract text.}
    \label{fig:Extraction-Prompt}
\end{figure*}

\paragraph{E2E recombination extraction} We use \texttt{Mistral-7B} as the backbone for our recombination extraction baseline. We fine-tune the model using \texttt{mistral-finetune}\footnote{https://github.com/mistralai/mistral-finetune} on a single NVIDIA RTX A6000 48GB GPU over $500$ steps. The training was conducted using the default learning rate of $6.e-5$ and weight decay of $0.1$. We use a batch size of $1$ and a maximum sequence length of $4096$ tokens. \texttt{mistral-finetune} implements Low-Rank Adaptation of LLM (LoRA), a parameter efficient fine-tuning method \cite{Hu2021LoRALA}, which we use with the default rank of $64$. The evaluation uses the corresponding repository, mistral-inference\footnote{https://github.com/mistralai/mistral-inference}. We rerun the same experiment using \texttt{Llama-3.1-8B} as a backbone, using an additional 500 warm-up steps, a learning rate of $2e-5$ and a weight decay of $0.01$. Figure \ref{fig:Extraction-Prompt} presents the prompt for these experiments.

\begin{figure}[!htb]
    \centering
    \includegraphics[width=\columnwidth]{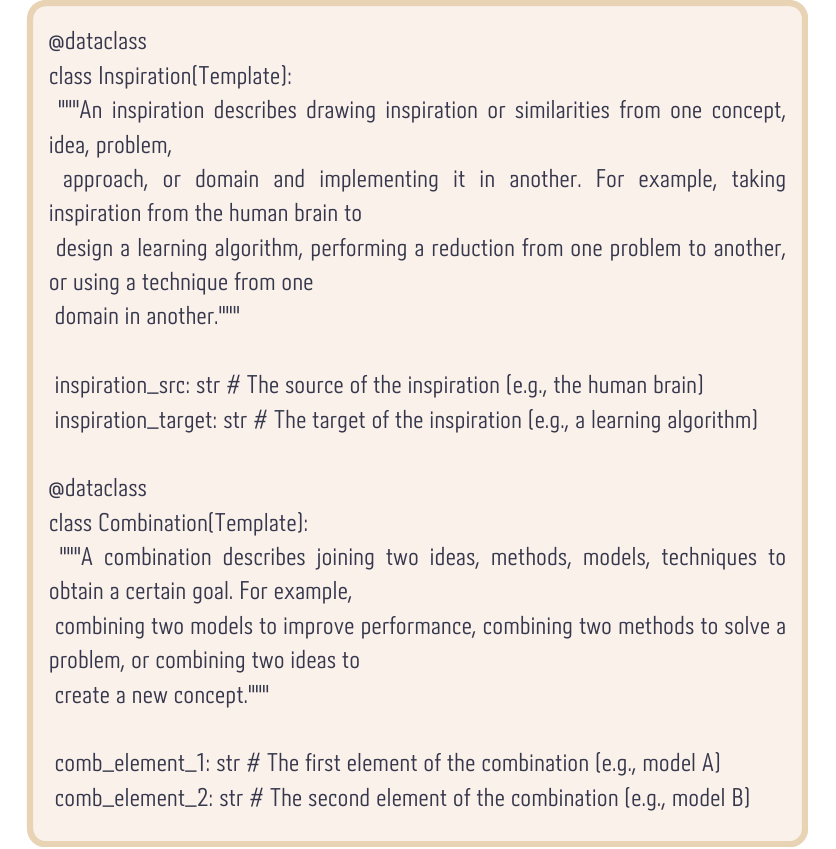}
    \caption{GoLLIE guidelines.}
    \label{fig:gollie_guidelines}
\end{figure}

In addition to fine-tuning LLMs on our data, we experiment with \texttt{GoLLIE} \cite{Sainz2023GoLLIEAG}, a general IE model fine-tuned to follow any annotation guidelines in a zero-shot fashion. We apply \texttt{GollIE-13B} on our data, using a single NVIDIA RTX A6000 48GB GPU, 1-beam search, and limit the new token number to 128. \texttt{GoLLIE} is finetuned from \texttt{CODE-LLaMA2}, and receives guidelines in the form of data classes describing what objects and properties the model should extract. Figure \ref{fig:gollie_guidelines} depicts the guidelines we used to test \texttt{GoLLIE} as an E2E recombination extraction model. In the rare cases where the model returns more than a single recombination type ($<10$), we select the first.

\begin{figure*}[!htbp]
    \centering
    \includegraphics[width=\textwidth]{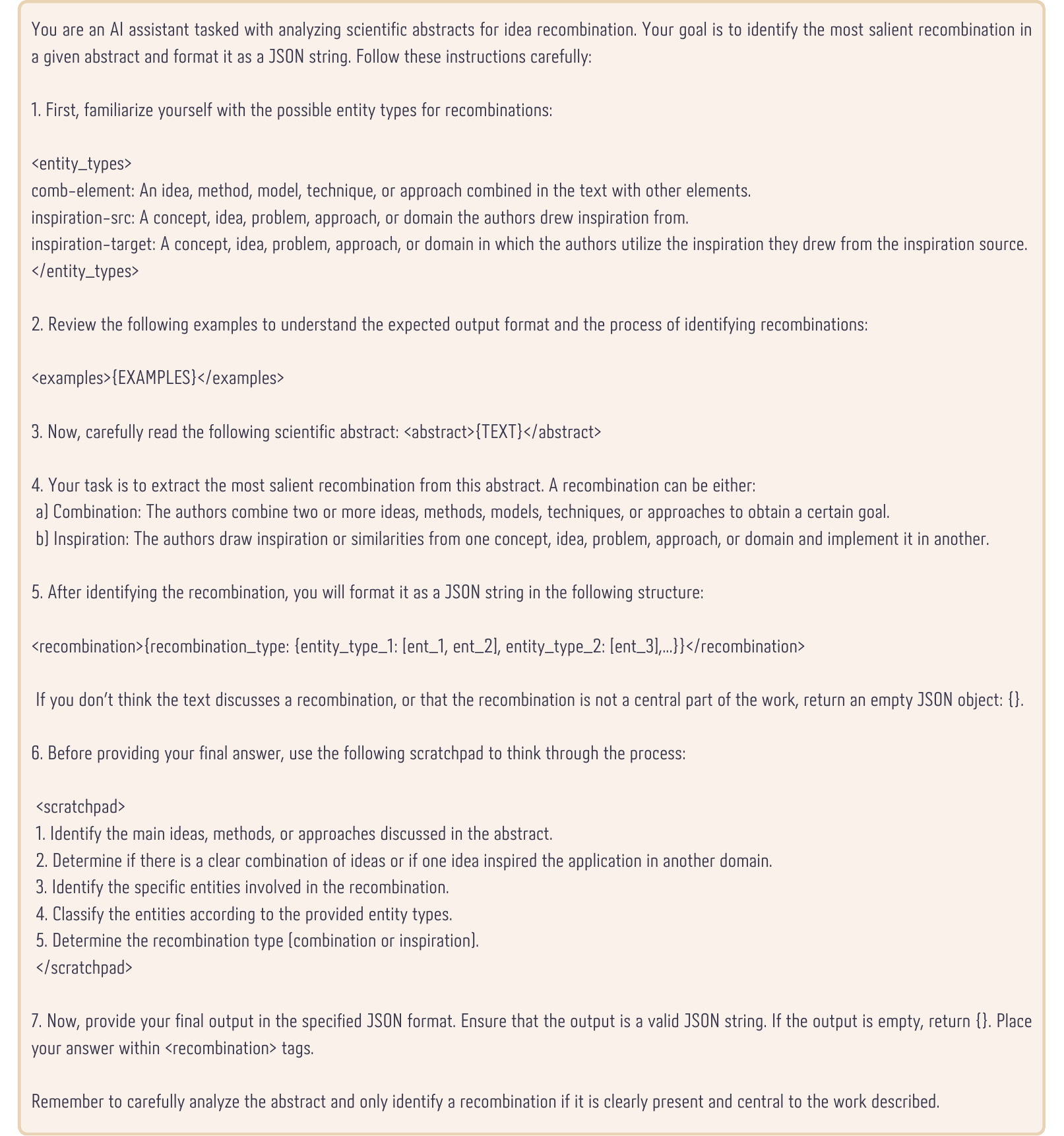}
    \caption{E2E ICL prompt. \{TEXT\} is a placeholder for the abstract text, and \{EXAMPLES\} for the ICL examples.}
    \label{fig:e2e_icl_prompt}
\end{figure*}

We also experiment with \texttt{GPT-4o} in few-shot settings. We select 45 examples for each example type (\textit{blend}, \textit{inspiration}, \textit{not-present}) from the training data (a total of 135). As Table \ref{table:chimera-s-splits} describes, the training set only has 45 \textit{inspiration} examples (as opposed to $> 100$ \textit{blend} and \textit{not-present} examples). $45$ is, therefore, the maximal number of examples per class we can sample while keeping the ICL set balanced. We run each experiment $5$ times, sampling a new set of few-shot examples in each, and report the average. Figure \ref{fig:e2e_icl_prompt} presents the prompt for this experiment.

\paragraph{Specialized baselines} The recombination extraction model has to execute multiple tasks at once (classifying the document, extracting entities, inferring relations), which might be more challenging than performing them separately. To explore this question, we examine our model classification and extraction abilities against designated models for each task.
We use \texttt{Mistral-7B} as a specialized classifier and experiment with two versions of the training data. The first includes binary responses (\textit{present}, \textit{not-present}), while the other contains a short CoT-style analysis string as well as the gold class. We construct the analysis string by incorporating the human entity annotations into predetermined templates (e.g., "\textit{This paper discusses a recombination since the authors take inspiration from [inspiration-source] and implement it in [inspiration-target]}").

\begin{figure}[!htb]
    \centering
    \includegraphics[width=\columnwidth]{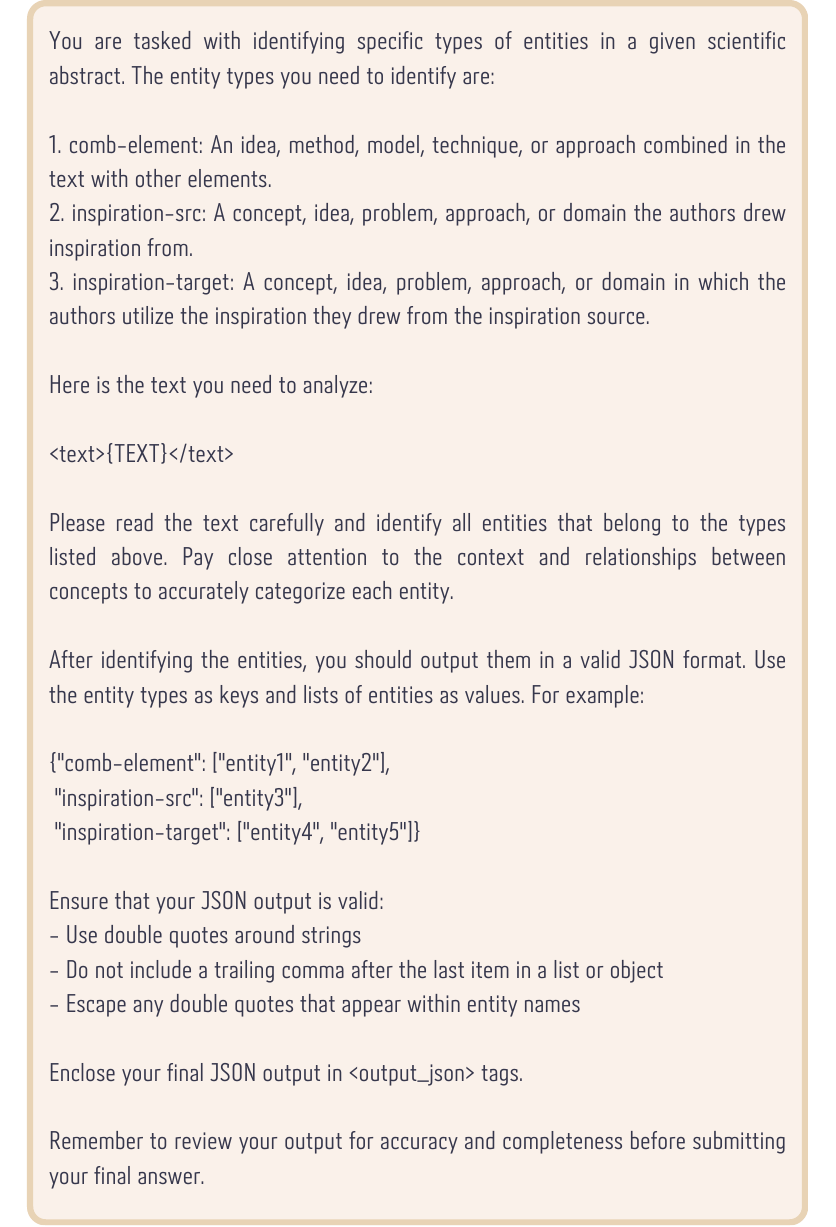}
    \caption{Entity extraction prompt. \{TEXT\} is a placeholder for the input abstract.}
    \label{fig:ner_icl_prompt}
\end{figure}

To evaluate entity extraction, we compare our model against \texttt{GPT-4o} in few-shot settings and include $45$ cases per example type, similarly to the E2E experiment. To account for variability due to example selection, we run each experiment $5$ times, sampling a new set of few-shot examples in each, and report the average. The total cost of this process sums up to 50\$. The prompt template for this experiment is available on Figure \ref{fig:ner_icl_prompt}.

We experiment with non-generative approaches as well, and compare our model to a \texttt{SciBERT} \cite{Zhong2021AFE} based token classifier. The encoder uses a standard Hugging-Face implementation of \texttt{SciBERT}, which we train on a single NVIDIA RTX A6000 48GB GPU over 500 steps. We use a weight decay of $0.1$, a learning rate of $6.e-5$ and a batch size of $1$. We also experiment with PURE \cite{Zhong2021AFE}, a well-known information extraction baseline. We finetune PURE over our train set using the default parameters, except for max\_span\_length, which we set to 40 to accommodate for the longer entities in our data.

\subsection{E2E vs Specialized extraction}\label{sec:e2e_vs_spacial}
This section reflects on the results described in Section \ref{sec:results}, drawing on implementation details of the baselines (described in Appendix \ref{sec:extraction_baselines}). In Section \ref{sec:results}, we observe that narrowing the focus to a smaller portion of the recombination extraction task does not always improve performance - in fact, it can lead to worse results. This pattern emerges across three Mistral-based classifiers: the end-to-end version (E2E), the specialized version (Abstract-classifier), and the specialized version trained with synthetic CoT strings (Abstract-classifier-CoT). We hypothesize that identifying recombination relations in text may be analogous to Chain-of-Thought prompting (CoT), a technique known to enhance LLM performance across various tasks \cite{wei2022chain}. This hypothesis is supported by the superior performance of Abstract-classifier-CoT compared to its non-CoT counterpart.

\subsection{Span similarity}\label{sec:span_sim_prompt}
\begin{figure}[!htb]
    \centering
    \includegraphics[width=\columnwidth]{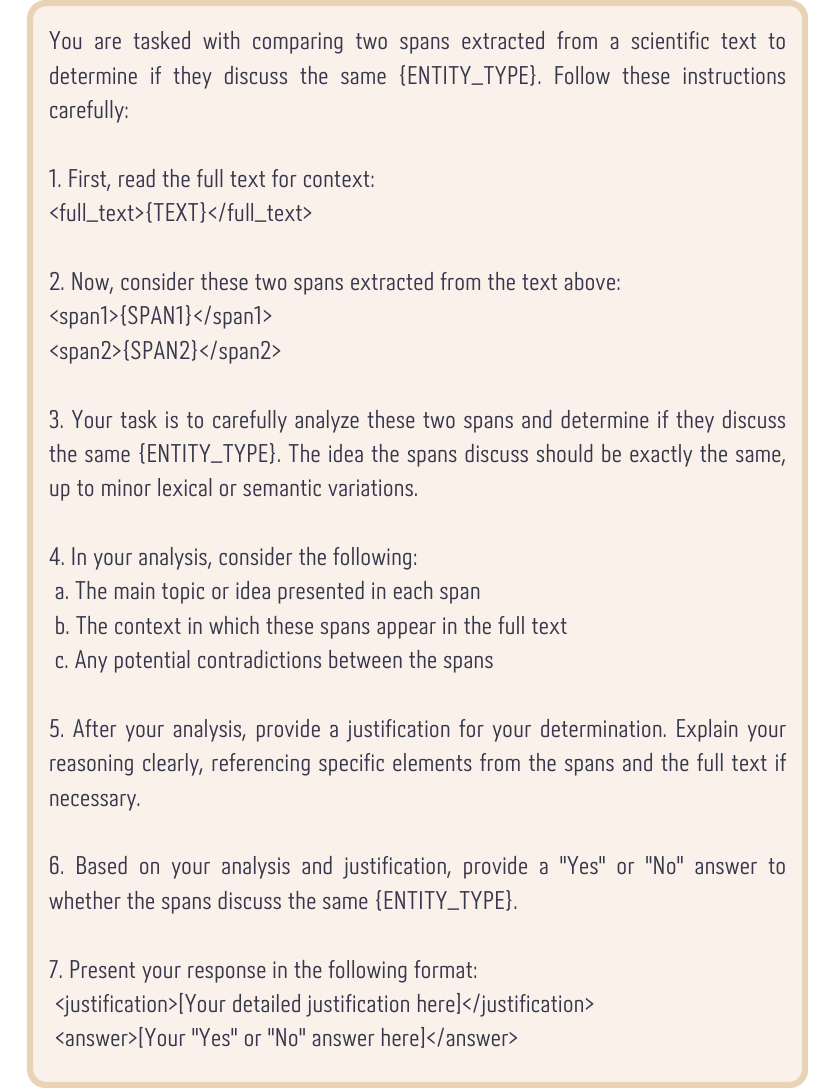}
    \caption{Span similarity prompt. \{ENTITY\_TYPE\} is either "combination-element", "inspiration-source" or "inspiration-target". \{TEXT\} is a placeholder for the paper's abstract. \{SPAN1\}, \{SPAN2\} are placeholders for the compared spans.}
    \label{fig:span_similarity_prompt}
\end{figure}

We provide our span similarity prompt in Figure \ref{fig:span_similarity_prompt}. We use it in the extraction evaluation process as discussed in Section \ref{eval_criteria}. To mitigate position bias, we query the model twice per pair with reversed orderings, accepting a match only if both judgments are positive. We prefer \texttt{GPT-4o-mini} over \texttt{GPT-4o} based on a comparison which found only $3$ disagreements across the test set.

\subsection{Extraction error analysis}\label{sec:qualitative_analysis}
 We perform analysis over the test set, revealing different sources of error which may inspire future improvements. Our focus is on understanding how different types of input texts can influence the result, specifically, in cases where the extraction model struggles. We use our best-performing fine-tuned E2E model for this analysis.

\begin{table*}[!htbp]
\centering
\footnotesize
\begin{tabular}{@{}p{15cm}@{}}
\toprule
\multicolumn{1}{c}{\textbf{Bad extraction examples (human annotated test set)}}\\
\midrule
\textbf{Abstract}: "...\marker{Kahneman \& Tversky’s prospect theory} tells us that humans perceive random variables in a biased but well-defined manner (1992) ... Using a Kahneman-Tversky model of human utility, we propose \marker{a HALO [Human Aware Loss Function]} that directly maximizes the utility of generations instead of maximizing the log-likelihood of preferences, as current methods do..." 
  \newline \newline
  \gr{Gold} = [\textbf{Inspiration}:  "\textit{Kahneman \& Tversky’s prospect theory}" $\longrightarrow$ "\textit{a HALO}"] \newline
  \rd{Pred} = []
  \\ \midrule
 
\textbf{Abstract}: "...We address the problem by proposing \marker{a Wasserstein GAN} combined with \marker{a new reverse mask operator, namely Reverse Masking Network (R-MNet)}, a perceptual adversarial network for image inpainting ... Additionally, we propose a new \marker{loss function} computed in feature space to target only valid pixels combined with \marker{adversarial training}..."
\newline \newline
  \gr{Gold} = [\textbf{Blend}: "\textit{a Wasserstein GAN}" $\longleftrightarrow$ "\textit{...R-MNet}"] \newline 
  \rd{Pred} = [\textbf{Blend}: "\textit{a Wasserstein GAN}" $\longleftrightarrow$ "\textit{...R-MNet}" $\longleftrightarrow$  "\textit{a new loss function}"]
\\ \midrule

\textbf{Abstract}: "... In order to characterize model flaws and choose a desirable representation, model builders often need to compare across multiple embedding spaces, a challenging analytical task supported by few existing tools. We first interviewed nine embedding experts in a variety of fields to characterize the diverse challenges they face and techniques they use when analyzing embedding spaces. Informed by these perspectives, we developed a novel system called Emblaze that integrates \marker{embedding space comparison} within a \marker{computational notebook environment}..."
\newline \newline
  \gr{Gold} = [\textbf{Blend}: "\textit{embedding space comparison}" $\longleftrightarrow$ "\textit{...notebook environment}"] \newline
  \rd{Pred} = []
\\ \bottomrule
\end{tabular}
\caption{In the first row, the extraction model misses an inspiration relation because of subtle phrasing. In the second row, when analyzing an abstract with multiple recombinations, the model fails to identify the most important one and confuses entities across different relations. In the third row, the model fails to detect a weak recombination example.}\label{table:error_analysis}%
\end{table*}

 \paragraph{Context dependent or subtle phrasing} We observe that, unsurprisingly, cases in which the recombination is implied or subtle are more challenging for the model. For instance (see also Table \ref{table:error_analysis}, row 1), \textit{"Kahneman \& Tversky’s prospect theory"} inspires the design of a loss function that \textit{"directly maximizes the utility of generations"}, but this is not stated directly. Moreover, abstracts that express idea recombination while referencing previously mentioned entities are also harder to detect. 
 \paragraph{Multiple recombinations} Some papers present a salient recombination along with other insignificant ones. We notice that in those cases, the model might extract a non-salient recombination or mix multiple ones (see Table \ref{table:error_analysis}, row 2 for such a case).  
  \paragraph{Borderline cases} The role of a recombination as a core element in the work is sometimes debatable. Table \ref{table:error_analysis}, row 3 presents an example of such a case where the authors explicitly mention integrating \textit{"embedding space comparison"} with \textit{"computational notebook environment"}, which may be interpreted as a recombination (the usage of notebook in these environments is completely new and novel), or simply as a way to present the tool's environment. We notice that the extraction model tends to miss those cases.

\begin{table*}[!htbp]
    \centering
    \footnotesize
    \begin{tabular}{p{15cm}}
         \toprule
         \multicolumn{1}{c}{\textbf{Bad Extraction Examples (arXiv)}}\\ \midrule
         \textbf{Abstract}: "The detection of allusive text reuse is particularly challenging due to the sparse evidence on which allusive references rely---commonly based on none or very few shared words. Arguably, lexical semantics can be resorted to since uncovering semantic relations between words has the potential to increase the support underlying the allusion and alleviate the lexical sparsity. A further obstacle is the lack of evaluation benchmark corpora, largely due to the highly interpretative character of the annotation process. \marker{In the present paper, we aim to elucidate the feasibility of automated allusion detection. We approach the matter from an Information Retrieval perspective in which referencing texts act as queries and referenced texts as relevant documents to be retrieved, and estimate the difficulty of benchmark corpus compilation by a novel inter-annotator agreement study on query segmentation}..."\newline\newline
         \textcolor{red}{\faTimes}\xspace\textbf{Automatic extraction} (incorrect entities): [\textbf{Inspiration}: "In an Information Retrieval perspective, referencing texts act as queries and referenced texts as relevant documents to be retrieved" $\longrightarrow$ \rd{"a task-oriented dialog system"}]\\ \midrule
         \textbf{Abstract}: "Supervised deep learning with pixel-wise training labels has great successes on multi-person part segmentation. However, data labeling at pixel-level is very expensive. To solve the problem, people have been exploring to use synthetic data...the results are much worse compared to those using real data and manual labeling. The degradation of the performance is mainly due to the domain gap, i.e., the discrepancy of the pixel value statistics between real and synthetic data. In this paper, we observe that \marker{real} and \marker{synthetic humans} both have a skeleton (pose) representation. We found that the skeletons can effectively bridge the synthetic and real domains during the training. Our proposed approach takes advantage of the rich and realistic variations of the real data and the easily obtainable labels of the synthetic data to learn multi-person part segmentation on real images without any human-annotated labels... "\newline\newline
         \textcolor{red}{\faTimes}\xspace\textbf{Automatic extraction} (uninformative entities): [\textbf{Combination}: \rd{"real"} $\longleftrightarrow$ \rd{"synthetic humans"}]\\ \midrule
         \textbf{Abstract}: "...In this paper we propose an LLM feature-based framework for dialogue constructiveness assessment that combines the strengths of \marker{feature-based} and \marker{neural} approaches, while mitigating their downsides. The framework first defines a set of dataset-independent and interpretable linguistic features, which can be extracted by both prompting an LLM and simple heuristics. Such features are then used to train LLM feature-based models...We also find that the LLM feature-based model learns more robust prediction rules instead of relying on superficial shortcuts, which often trouble neural models."\newline\newline
         \textcolor{red}{\faTimes}\xspace\textbf{Automatic extraction} (uninformative entities): [\textbf{Combination}: \rd{"neural"} $\longleftrightarrow$ \rd{"feature-based"}]\\
         \bottomrule
    \end{tabular}
    \caption{\revision{Representative examples of bad automatic extraction. Many errors stem from uninformative entity spans, as presented by the two bottom examples.}}\label{tab:automatic_extraction}
\end{table*}

\subsection{Large-scale extraction assessment}\label{section:qualitative_extraction_eval}
\begin{figure}[!htb]
    \centering
    \includegraphics[width=0.95\columnwidth]{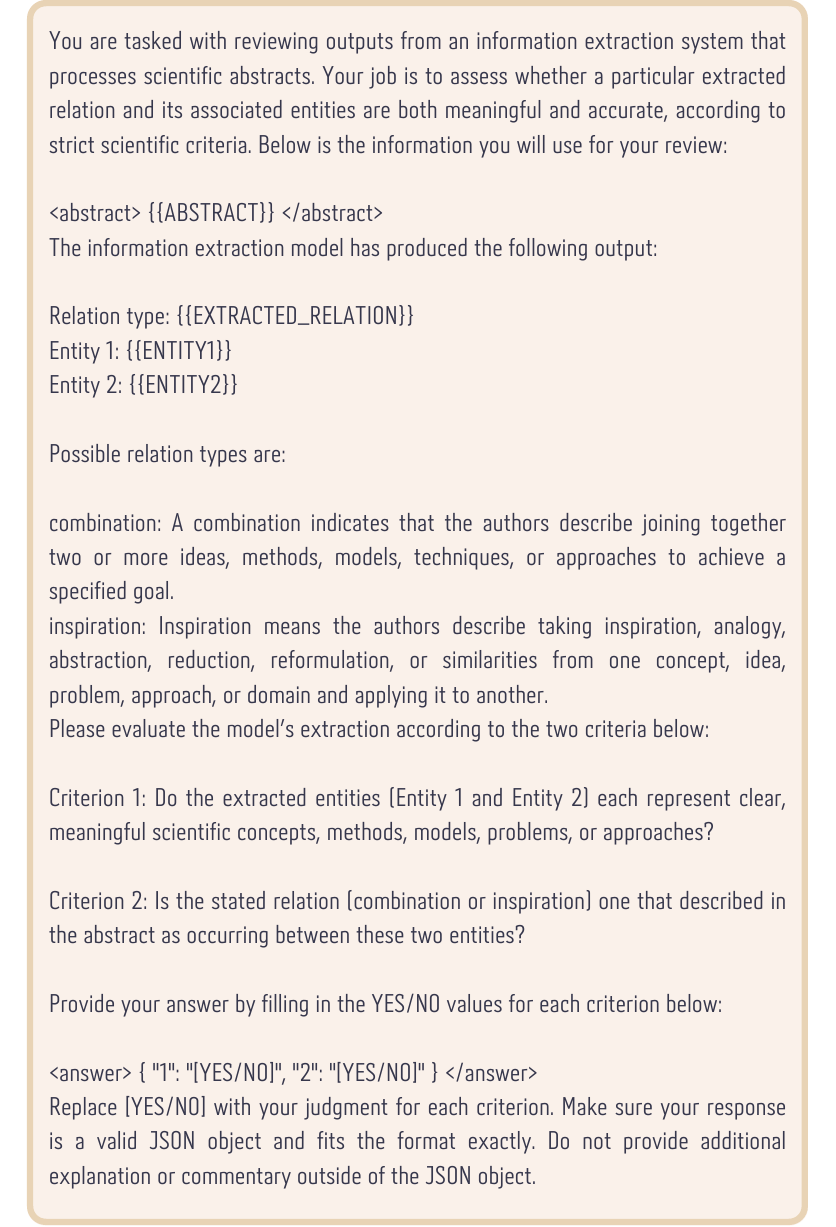}
    \caption{\revision{Large-scale evaluation prompt. \{ABSTRACT\} is a placeholder for the original abstract text. \{EXTRACTED\_RELATION\}, \{ENTITY1\}, and \{ENTITY2\} are placeholders for the relation type and entities extracted by our model.}}
    \label{fig:large_scale_prompt}
\end{figure}

\revision{To complement our human annotation efforts and enable large-scale evaluation, we conducted a qualitative assessment of the automatically extracted recombination examples in \kbname using GPT-4.1 as an LLM-based judge.}

\paragraph{Validating the LLM Judge.}
\revision{We first assessed GPT-4.1’s reliability by comparing its judgments against those of a domain expert. A PhD student with NLP expertise manually reviewed $100$ randomly sampled recombination examples and labeled each as correct if: (1) the extracted entities corresponded to meaningful scientific concepts, and (2) the relation between them captured a central recombination explicitly described in the abstract. Upon analyzing the identified extraction errors, we observe a significant portion stems from extracting correct recombinations with uninformative entities (criteria 2) and not from a conceptual misunderstanding of the text. We provide examples of such cases in Table \ref{tab:automatic_extraction}.}

\revision{GPT-4.1 was prompted with the same examples using an evaluation template aligned with the assessment criteria (see Figure \ref{fig:large_scale_prompt}). Given the imbalance nature of the data, we report the F1 score instead of Cohen's $\kappa$, following the recommendations of previous work \cite{delgado2019cohen}. The resulting F1 score of $0.912$ indicates substantial agreement, supporting the use of GPT-4.1 as a reliable proxy for large-scale quality assessment.}

\paragraph{Large-Scale Evaluation.}
\revision{Following validation, we applied GPT-4.1 to a larger sample of 2,000 automatically extracted examples from \kbname. The model labeled 799 of these examples as correct, resulting in an estimated extraction accuracy of 80.55\%. These results provide further evidence for the overall quality and robustness of our extraction pipeline.}

\section{Additional KG Construction Details}\label{sec:kg_const}
\subsection{Graph nodes domains}\label{sec:domain_anlysis}
\begin{table*}[!htbp]
\centering
 \footnotesize
\begin{tabular}{@{}lll@{}}
\toprule
\multicolumn{3}{c}{\textbf{Non-arXiv scientific domains}}\\
\midrule
Agricultural Science   & Anatomy               & Animal Science         \\
Anthropology           & Archaeology           & Behavioral Science     \\
Biochemistry           & Bioinformatics        & Bioclimatology         \\
Biomedical Engineering & Biophysics            & Biotechnology          \\
Botany                 & Cardiology            & Chemical Engineering   \\
Civil Engineering      & Clinical Psychology   & Cognitive Science      \\
Criminology            & Cryosphere Science    & Cytology               \\
Demography             & Dentistry             & Dermatology            \\
Developmental Biology  & Ecology               & Ecotoxicology          \\
Economics              & Educational Psychology & Electrical Engineering \\
Emergency Medicine     & Endocrinology         & Energy Science         \\
Engineering Science    & Entomology            & Environmental Engineering \\
Environmental Science  & Epidemiology          & Ethology               \\
Food Science           & Forestry              & Gastroenterology       \\
Genetics               & Genomics              & Geography              \\
Geology                & Geophysics            & Glaciology             \\
Health Informatics     & Histopathology        & Hydrodynamics          \\
Hydrogeology           & Hydrology             & Immunogenetics         \\
Immunology             & Industrial/Organizational Psychology & Landscape Architecture \\
Linguistics            & Marine Biology        & Materials Science      \\
Mechanical Engineering & Medical Microbiology  & Meteorology            \\
Microbiology           & Mineralogy            & Molecular Biology      \\
Mycology               & Nanotechnology        & Neurology              \\
Neuroscience           & Nuclear Engineering   & Nutritional Science    \\
Obstetrics             & Oceanography          & Oncology               \\
Ophthalmology          & Ornithology           & Orthopedics            \\
Otology                & Paleoclimatology      & Paleontology           \\
Pathobiology           & Pathology             & Pediatric Medicine     \\
Pedagogy               & Petrology             & Pharmacogenomics       \\
Pharmacology           & Philosophy            & Physiology             \\
Political Science      & Proteomics            & Psychiatry             \\
Psychology             & Psychopathology       & Public Health          \\
Pulmonology            & Radiology             & Rheumatology           \\
Seismology             & Social Psychology     & Sociology              \\
Surgery                & Systems Biology       & Thermodynamics         \\
Toxicology             & Urban Planning        & Urology                \\
Veterinary Science     & Virology              & Volcanology            \\
Wildlife Biology       & Zoology               &             \\
\bottomrule
\end{tabular}
\caption{Non-arXiv scientific domains. We complement arXiv category taxonomy using a broader list of scientific fields.}\label{table:sci-branches}
\end{table*}

\begin{figure*}[!htbp]
    \centering
    \includegraphics[width=\textwidth]{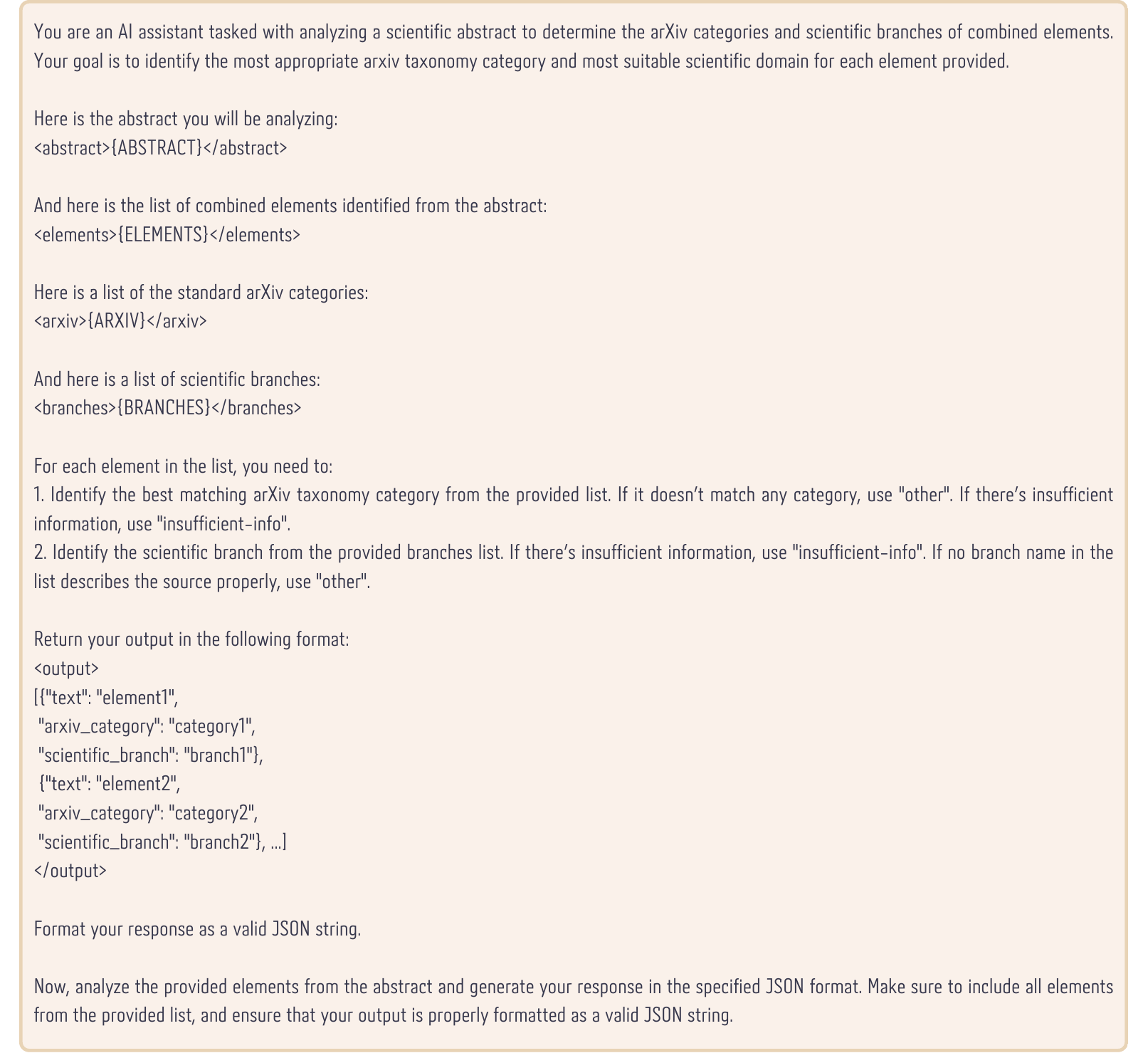}
    \caption{blend domain analysis prompt. \{ELEMENTS\} is a placeholder for the recombination entities extracted from \{ABSTRACT\}. \{ARXIV\} is a placeholder for full arXiv category names and their descriptions. \{BRANCHES\} is a placeholder for the list of non-arXiv domains given in Appendix \ref{sec:domain_anlysis}, Table \ref{table:sci-branches}.}
    \label{fig:comb_domain_prompt}
\end{figure*}

\begin{figure*}[!htbp]
    \centering
    \includegraphics[width=\textwidth]{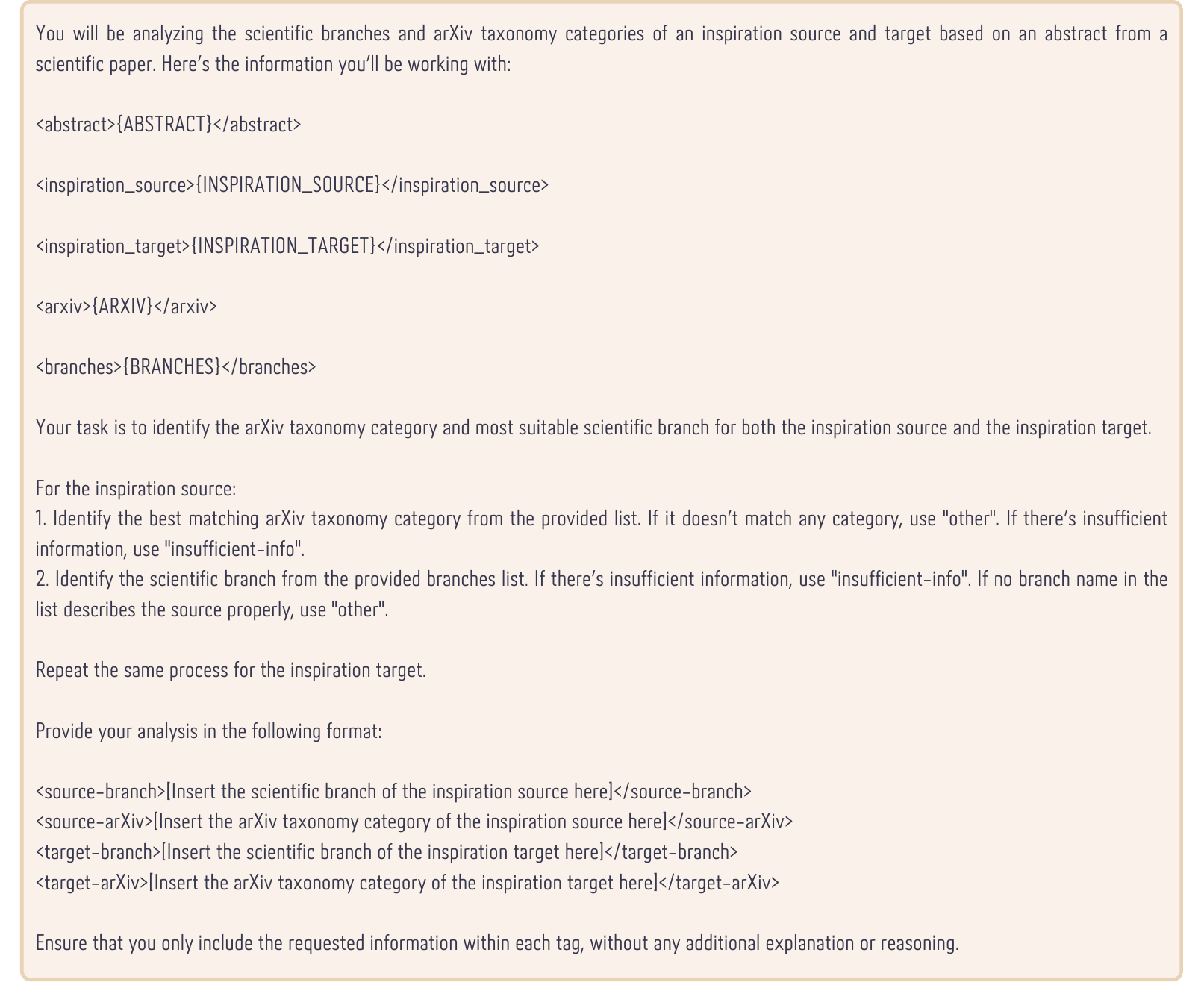}
    \caption{inspiration domain analysis prompt. \{INSPIRATION\_SOURCE\} and \{INSPIRATION\_TARGET\} are placeholders for the inspiration entities extracted from \{ABSTRACT\}. \{ARXIV\} is a placeholder for full arXiv category names and their descriptions. \{BRANCHES\} is a placeholder for the list of non-arXiv domains given in Appendix \ref{sec:domain_anlysis}, Table \ref{table:sci-branches}.}
    \label{fig:inspo_domain_prompt}
\end{figure*}

We identify the scientific domain of each entity using \texttt{GPT-4o} in a zero-shot setting. Given the abstract and the extracted recombination entities, the model assigns to each entity an \textit{arXiv category} and a broader \textit{scientific branch}. If the model successfully assigns an arXiv category, we treat it as the entity’s domain.

Otherwise, the model selects a branch from a predefined list of outer-arXiv domains (see Table~\ref{table:sci-branches}) and sets it as the domain. If neither a standard arXiv category nor a branch can be assigned, the entity is labeled as belonging to the \texttt{Other} domain.

\camera{Entities labeled as \texttt{Other} are excluded from the analysis in Section~\ref{subsec2:kb_analysis}, to avoid interpreting overly broad or miscellaneous signals, though they are fully preserved within the predictive experiments.}
Figures~\ref{fig:comb_domain_prompt} and~\ref{fig:inspo_domain_prompt} present the prompts used for analyzing blend and inspiration relations, respectively. Running this classification process over the full corpus costs approximately 250\$.

\begin{table*}[!htbp]
\centering
\footnotesize
\begin{tabular}{@{}lp{13cm}@{}}
\toprule
\textbf{Type} & \textbf{Examples} \\
\midrule
Non-Academic & "the snap-through action of a steel hairclip", "yoga", \newline "origami, the traditional Japanese paper-folding technique, is a powerful metaphor for design and fabrication of reconfigurable structures", "Tangram, a game that requires replicating an abstract pattern from seven dissected shapes" \\
\midrule
Noisy & "a deep", "word-", "at the context level", "a neural part", "post", "text--audio", "end-to-end multi-modal model only X-VLM only X-VLM only X-VLM only X-VLM only X-VLM only X-VLM only X-VLM only X-VLM only X-VLMs",
 "a user's long-term"\\
\midrule
Overly-general & "human experiences", "a styling method", "local search method", "a pipeline inspired by experts' work", "a new modality", "feature based approaches"\\
\midrule
Misclassified & "Reinforcement learning, or RL", "Facial Expressions Recognition(FER)", "a Kullback-Liebler regularization function", "K-nearest neighbors algorithm", "Shapley values from game theory", "Gaussian Stochastic Weight Averaging"\\
\bottomrule
\end{tabular}
\caption{Examples of graph nodes in the "other" domain. We analyze a sample of 150 nodes in this domain and identify groups with common traits, as shown in the table.}\label{table:other_nodes}%
\end{table*}

\paragraph{The \texttt{Other} domain}
We assign the \texttt{Other} domain to nodes the model fails to classify. In total, 2,127 graph nodes fall into this category. We manually examined a sample of 150 such nodes and found that many were either too ambiguous or too general to categorize meaningfully. Interestingly, some of these nodes refer to non-academic or highly niche concepts (see examples in Table~\ref{table:other_nodes}).

\begin{table*}[!htbp]
\centering
\footnotesize
\begin{tabular}{@{}lp{11cm}@{}}
\toprule
\textbf{Group} & \textbf{Scientific domains} \\
\midrule
Geosciences & Geology, Geophysics, Petrology, Mineralogy, Hydrology, Hydrogeology, Seismology, Volcanology, Cryosphere Science, Glaciology, Geography \\ \hline
Environmental Sciences & Environmental Science, Environmental Engineering, Ecology, Ecotoxicology \\ \hline
Biomedical Sciences & Biochemistry, Immunology, Immunogenetics, Neuroscience, Oncology, Pathology, Pathobiology, Pharmacology, Toxicology \\ \hline
Health and Medicine & Cardiology, Neurology, Urology, Gastroenterology, Obstetrics, Pediatric Medicine, Rheumatology, Dermatology, Ophthalmology, Otology, Pulmonology, Emergency Medicine, Surgery, Radiology, Orthopedics, Psychiatry, Dentistry, Public Health, Epidemiology, Health Informatics, Clinical Psychology, Psychopathology \\ \hline
Zoology & Zoology, Entomology, Ornithology, Wildlife Biology, Animal Science, Veterinary Science, Ethology \\ \hline
Agriculture & Agricultural Science, Forestry \\ \hline
Food Sciences & Nutritional Science, Food Science \\ \hline
Psychology & Educational Psychology, Social Psychology, Psychology, Industrial/Organizational Psychology \\ \hline
Microbiology & Microbiology, Medical Microbiology \\ \hline
Humanities & Linguistics, Philosophy, Pedagogy \\ \hline
Social Sciences & Sociology, Anthropology, Political Science, Demography \\
\bottomrule
\end{tabular}
\caption{Scientific domains grouped by category. We group similar non-arXiv scientific domains (see Table \ref{table:sci-branches}) to thicken infrequent ones.}\label{tab:domain_groups}
\end{table*}

\begin{table*}[!htbp]
\centering
\footnotesize
\begin{tabular}{@{}lrlrlr@{}}
\toprule
\textbf{Domain} & \textbf{Count} & \textbf{Domain} & \textbf{Count} & \textbf{Domain} & \textbf{Count} \\
\midrule
cs.cv & 12504 & cs.lg & 8440 & cs.cl & 4697 \\
cs.ro & 2241 & cs.ai & 2091 & cognitive science & 936 \\
cs.ir & 884 & cs.ne & 864 & cs.si & 655 \\
cs.hc & 645 & q-bio.nc & 441 & cs.ds & 409 \\
cs.cg & 382 & cs.cy & 378 & cs.gr & 367 \\
math.oc & 356 & eess.iv & 278 & cs.dm & 269 \\
cs.db & 254 & eess.sp & 242 & cs.lo & 204 \\
cs.ma & 203 & cs.ce & 185 & cs.sy & 177 \\
cs.cr & 164 & stat.me & 138 & cs.gt & 132 \\
psychology & 116 & eess.sy & 108 & cs.se & 104 \\
zoology & 101 & cs.it & 100 & math.pr & 96 \\
cs.dc & 89 & behavioral science & 88 & cs.mm & 82 \\
eess.as & 79 & nlin.ao & 79 & cs.ar & 74 \\
cs.na & 66 & cs.pl & 65 & biomedical sciences & 63 \\
physics.med-ph & 60 & stat.ml & 56 & health and medicine & 56 \\
physics.bio-ph & 52 & cs.ni & 48 & physics.ao-ph & 44 \\
stat.th & 43 & anatomy & 41 & math.na & 40 \\
math.ds & 39 & cs.fl & 38 & humanities & 38 \\
q-bio.pe & 32 & cs.dl & 32 & cs.sc & 30 \\
math-ph & 27 & cond-mat.stat-mech & 25 & math.ap & 24 \\
math.dg & 22 & physics.class-ph & 22 & cs.sd & 22 \\
econ.th & 21 & math.ca & 21 & math.mg & 20 \\
physics.comp-ph & 20 & physics.optics & 20 & cs.et & 20 \\
\bottomrule
\end{tabular}
\caption{Node domains distribution. The table presents the number of graph nodes from each domain with above-median frequency.}\label{table:node_domain_distribution}%
\end{table*}

\paragraph{domain grouping} To avoid sparsity, we group similar domains as displayed in Table \ref{tab:domain_groups}. Table \ref{table:node_domain_distribution} presents the node distribution of common domains after applying this grouping process.

\subsection{Entity Normalization} \label{sec:entity_norm}
\camera{Our entity normalization process consists of the following steps:}

\begin{enumerate}
    \item \camera{\textbf{Abbreviation Expansion}: We standardize entities using (a) Regex-based cleaning (e.g., ``\textit{Chain of Thought (CoT)}'' $\rightarrow$ ``\textit{Chain of Thought}'') and (b) Scispacy's abbreviation expansion mechanism\footnote{\url{https://github.com/allenai/scispacy\#abbreviationdetector}}, which resolves short-forms to their long-form versions using the document context (e.g., ``\textit{CoT}'' $\rightarrow$ ``\textit{Chain of Thought}'').}
    
    \item \camera{\textbf{Agglomerative Clustering}: We group entities by performing agglomerative clustering with cosine similarity on \texttt{all-mpnet-base-v2} embeddings, using average linkage and a distance threshold of $0.05$.}
\end{enumerate}

\subsection{Entity Postprocessing}\label{sec:postprocessing}
\begin{figure*}[!htbp]
    \centering
    \includegraphics[width=\textwidth]{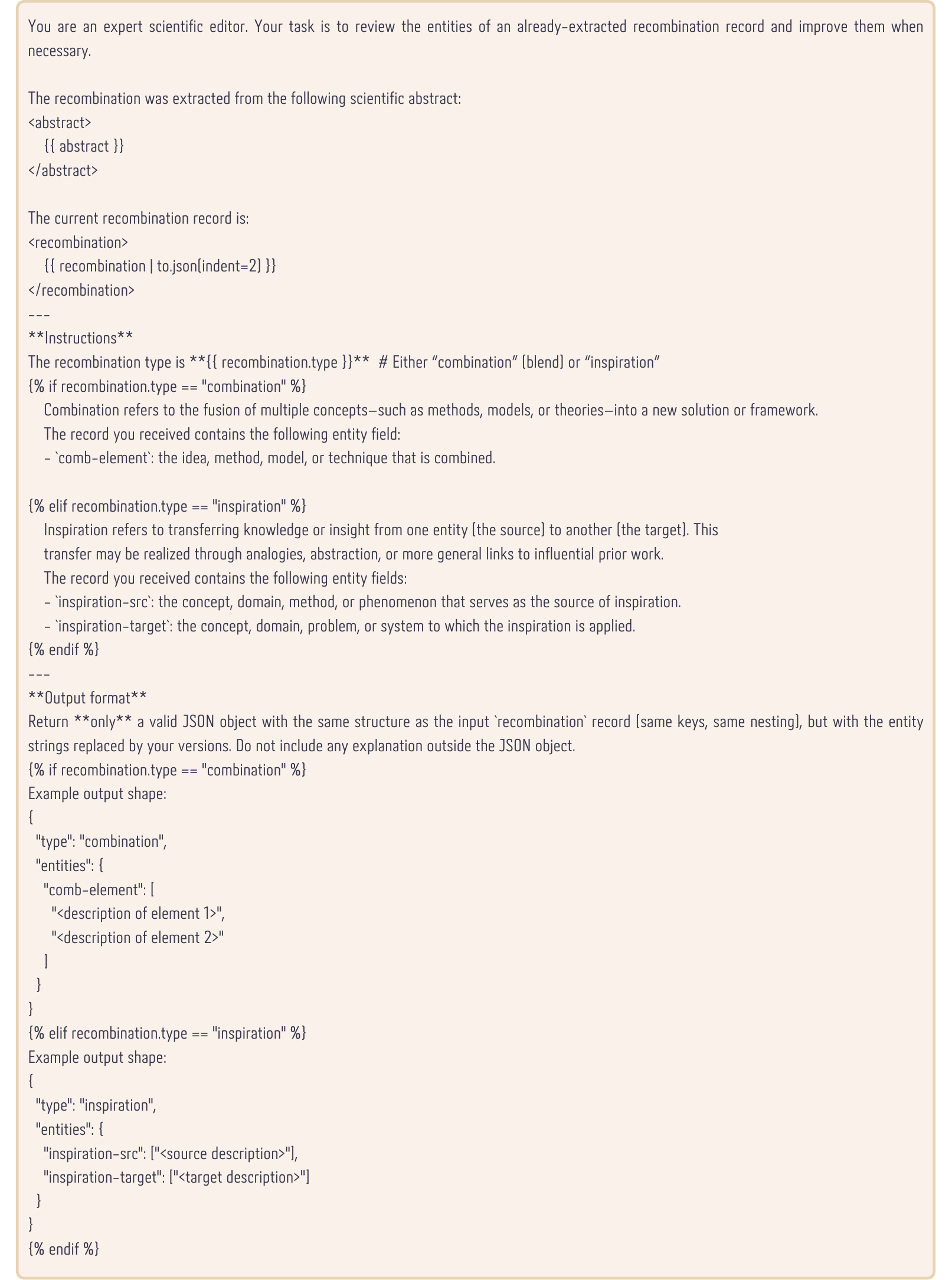}
    \caption{\camera{Entity postprocessing prompt. \{\{ABSTRACT\}\} is a placeholder for the source abstract. \{\{RECOMBINATION\}\} is a JSON object containing the extracted recombination record. The prompt conditionally adapts its instructions and output schema based on the recombination type (\textit{combination} or \textit{inspiration}).}}
    \label{fig:postprocessing_prompt}
\end{figure*}

\camera{After the initial large-scale extraction, we apply an LLM-based postprocessing step to refine the extracted entity spans. Given the source abstract and the extracted recombination record, a high-performance LLM (\texttt{claude-opus-4-6}) is prompted to review and improve the entity strings if necessary. Figure~\ref{fig:postprocessing_prompt} presents the prompt used for this step.}

\section{Additional Knowledge Base Analysis}\label{sec:graph_analysis}

\subsection{Keywords in Annotated Data}\label{sec:keyword_freq}
\camera{Table~\ref{table:keyword_freq} reports the $30$ most frequent keywords in the knowledge base. Importantly, keyword filtering was used solely to guide the annotation stage; the large-scale KB extraction itself was performed without any keyword restrictions. Of the extracted recombinations, $3592$ (approximately $13$\%) come from abstracts containing none of the keywords in our list, demonstrating that the extracted KB generalizes beyond the seed keywords. We provide examples of such recombinations in Table~\ref{table:no_keyword_examples}.} 

\begin{table}[!htb]
\centering
\footnotesize
\begin{tabular}{@{}lc@{\hskip 1.2em}lc@{}}
\toprule
\textbf{Keyword} & \textbf{\#} & \textbf{Keyword} & \textbf{\#} \\
\midrule
combine & 5,196 & fuse & 1,271 \\
relation & 3,756 & alignment & 1,266 \\
inspire & 3,753 & integrating & 1,260 \\
inspired & 3,655 & hybrid & 1,234 \\
integrate & 3,300 & merge & 1,227 \\
fusion & 3,072 & combined & 1,170 \\
view & 2,935 & correlation & 1,146 \\
combines & 2,885 & perception & 1,049 \\
align & 2,382 & mix & 988 \\
incorporate & 2,375 & cast & 972 \\
join & 2,260 & compose & 939 \\
combining & 2,046 & integrated & 890 \\
relate & 1,798 & associate & 839 \\
combination & 1,571 & synthesis & 780 \\
integrates & 1,496 & link & 733 \\

\bottomrule
\end{tabular}
\caption{Top-$30$ most frequent keywords in \kbname. \# is the number of abstracts containing the keyword.}\label{table:keyword_freq}
\end{table}

\begin{table*}[!htbp]
    \centering
    \footnotesize
    \begin{tabular}{p{15cm}}
    \toprule
    \multicolumn{1}{c}{\textbf{Recombination examples from abstracts with no seed keywords}} \\
    \midrule
    \textbf{Abstract}: ``...Recent \marker{white-box retouching} methods rely on cascaded global filters that provide image-level filter arguments but cannot perform fine-grained retouching. In contrast, \marker{colorists typically employ a divide-and-conquer approach, performing a series of region-specific fine-grained enhancements when using traditional tools like Davinci Resolve}. \bmarker{We draw on this insight} to develop a white-box framework for photo retouching using parallel region-specific filters...'' \newline

    \textbf{Inspiration}: ``\textit{colorists typically employ a divide-and-conquer approach, performing a series of region-specific fine-grained enhancements when using traditional tools like Davinci Resolve}'' $\longrightarrow$ ``\textit{white-box retouching}'' \\
    \midrule
    \textbf{Abstract}: ``\ldots \marker{Biological muscle movement consists of stretching and shrinking fibres via spike-commanded signals that come from motor neurons, which in turn are connected to a central pattern generator neural structure}. \ldots\ The ED-Scorbot platform \ldots\ implements \marker{a Spiking Proportional-Integrative-Derivative algorithm}, \bmarker{mimicking in this way the previously commented biological systems}\ldots'' \newline

    \textbf{Inspiration}: ``\textit{Biological muscle movement consists of stretching and shrinking fibres via spike-commanded signals that come from motor neurons, which in turn are connected to a central pattern generator neural structure}'' $\longrightarrow$ ``\textit{a Spiking Proportional-Integrative-Derivative algorithm}'' \\
    \bottomrule
    \end{tabular}
    \caption{\camera{Examples of recombinations extracted from abstracts containing none of the seed keywords (Table \ref{table:keywords}), demonstrating generalization of the extraction pipeline beyond keyword-filtered content. \sethlcolor{peach}\hl{Peach} highlights mark extracted entities; \bmarker{blue} highlights mark the connective expression (generalized beyond our initial keyword list) linking them.}}\label{table:no_keyword_examples}
\end{table*}

\subsection{Predominant recombination relations}\label{sec:predominant_recomb}
We provide a tabular version of Figure \ref{fig:frequent_recomb_sunkey} in Section \ref{subsec2:kb_analysis} on Table \ref{table:all_inspirations} for better readability.

\begin{table*}[!htbp]
    \centering
    \footnotesize

    \begin{tabular}{@{}llc|llc@{}}
        \toprule
        \multicolumn{3}{c|}{\textit{Inspirations}}& \multicolumn{3}{c}{\textit{Blends}}\\
        \midrule
        \textbf{Source} & \textbf{Target} & \textbf{Count}& \textbf{Source} & \textbf{Target} & \textbf{Count}\\
        \midrule
        cs.cv & cs.cv & 334 & cs.cv & cs.cv & 4329 \\
        cs.lg & cs.cv & 300 & cs.lg & cs.lg & 2793 \\
        cognitive science & cs.cv & 278 & cs.cl & cs.cl & 1049 \\
        cs.lg & cs.lg & 254 & cs.lg & cs.cv & 992 \\
        cognitive science & cs.lg & 211 & cs.cv & cs.lg & 470 \\
        cs.cl & cs.cl & 190 & cs.cl & cs.cv & 422 \\
        cs.cl & cs.cv & 188 & cs.cv & cs.cl & 391 \\
        cognitive science & cs.ai & 184 & cs.lg & cs.cl & 363 \\
        cognitive science & cs.cl & 142 & cs.ro & cs.ro & 299 \\
        cs.cl & cs.ai & 141 & cs.ro & cs.cv & 218 \\
        cs.lg & cs.ai & 118 & cs.cl & cs.lg & 197 \\
        q-bio.nc & cs.cv & 114 & cs.ai & cs.cl & 174 \\
        q-bio.nc & cs.lg & 102 & cs.ai & cs.ai & 161 \\
        cognitive science & cs.ro & 100 & cs.ai & cs.lg & 151 \\
        cs.cv & cs.lg & 94 & cs.lg & cs.ai & 146 \\
        cs.lg & cs.cl & 84 & cs.lg & cs.ne & 133 \\
        cs.cl & cs.lg & 84 & cs.ir & cs.ir & 132 \\
        math.oc & cs.lg & 83 & cs.lg & cs.ro & 124 \\
        zoology & cs.ro & 76 \\
        \bottomrule
    \end{tabular}
    \caption{Predominant inspiration and blend relations. The above is a tabular version of Figures \ref{fig:bleds_sunkey}, \ref{fig:inspiration_sunkey} in Section \ref{subsec2:kb_analysis}. It presents edges with (source-domain, target-domain) pairs frequency above the 0.98 quantile.}\label{table:all_inspirations}%
\end{table*}

\subsection{Nuanced recombination types}\label{sec:fine_grained_types}
\begin{table*}[!htbp]
    \centering
    \footnotesize
    \begin{tabular}{p{15cm}}
    \toprule 
    \multicolumn{1}{c}{\textbf{Nuanced recombination types examples}} \\
    \toprule

    \textbf{Abstract}: "Register allocation is one of the most important problems for modern compilers...This work demonstrates the use of casting the \marker{register allocation problem} as a \marker{graph coloring problem}..."\newline\newline
    \textbf{Inspiration}: "\textit{a graph coloring problem}" $\longrightarrow$ "\textit{Register allocation}" \\

    \midrule
    \textbf{Abstract}: "Affective sharing within groups strengthens coordination and empathy...we propose HeartBees, a bio-feedback system for visualizing collective emotional states, which maps \marker{a multi-dimensional emotion model} into a metaphorical visualization of \marker{flocks of birds}..." \newline\newline
    \textbf{Inspiration}: "\textit{flocks of birds}" $\longrightarrow$ "\textit{a multi-dimensional emotion model}" \\

    \midrule
    \textbf{Abstract}: "Physics-informed Graph Neural Networks have achieved remarkable performance...by mitigating common GNN challenges...Despite these advancements, the development of a simple yet effective paradigm that appropriately integrates previous methods for handling all these challenges is still underway. In this paper, we draw an analogy between \marker{the propagation of GNNs} and \marker{particle systems in physics}, proposing a model-agnostic enhancement framework..." \newline\newline
    \textbf{Inspiration}: "\textit{particle systems in physics}" $\longrightarrow$ "\textit{the propagation of GNNs}" \\
    \bottomrule
    \end{tabular}
    \caption{\revision{Examples of nuanced inspiration types found within \kbname. While all examples are labeled as \emph{inspiration}, they illustrate finer-grained mechanisms such as metaphor, reduction, and analogy. This suggests that our taxonomy is expressive enough to capture a rich diversity of recombination strategies.}}\label{tab:nuanced_recomb_types}
\end{table*}

\revision{In Section~\ref{methods:problem_definition}, we defined the two broad recombination types used in \kbname: \emph{blends} and \emph{inspirations}. In this section, we demonstrate that this taxonomy is both robust and expressive, offering broad coverage of more nuanced recombination phenomena.}

\revision{To support this, we perform a qualitative analysis of $30$ \emph{inspiration} examples from the \kbname dataset. We identify distinct subtypes of inspiration, such as \textit{analogy}, \textit{metaphor}, \textit{reduction}, \textit{abstraction}, and \textit{application of existing knowledge}. These subtypes emerge naturally within our current schema, illustrating its extensibility and broad coverage. Table \ref{tab:nuanced_recomb_types} provides examples for each. This analysis lays the groundwork for future refinement and expansion of the taxonomy.}

\section{Additional Prediction Details}\label{sec:pred_details}

\subsection{Prediction data preprocessing}\label{sec:pred_data_prep}
\begin{table*}[!htbp]
\centering
\footnotesize
\footnotesize
\begin{tabular}{@{}p{12cm}p{3cm}@{}}
\toprule
\textbf{Query} & \textbf{Answer} \\
\midrule
\textbf{Understanding the human brain's processing capabilities can inspire advancements in machine learning algorithms and architectures.} Previous methods in brain research were limited to identifying regions of interest for one subject at a time, restricting their applicability and scalability across multiple subjects.\newline\newline What would be a good source of inspiration for "\textbf{a highly efficient processing unit}"?& The human brain\\
\midrule
Existing models for link prediction in knowledge graphs primarily focus on \textbf{representing triplets in either distance or semantic space}, which limits their ability to fully capture the information of head and tail entities and utilize hierarchical level information effectively. \textbf{This indicates a need for improved methods that can leverage both types of information} for better representation learning in knowledge graphs.\newline\newline What could we blend with "\textbf{distance measurement space}" to address the described settings?& Semantic measurement space\\ 
\bottomrule
\end{tabular}
\caption{Leakages examples. Examples of leaks - queries that reveal or strongly imply the answer.}\label{table:leak_examples}%
\end{table*}

\begin{figure*}[!htbp]
    \centering
    \includegraphics[width=\textwidth]{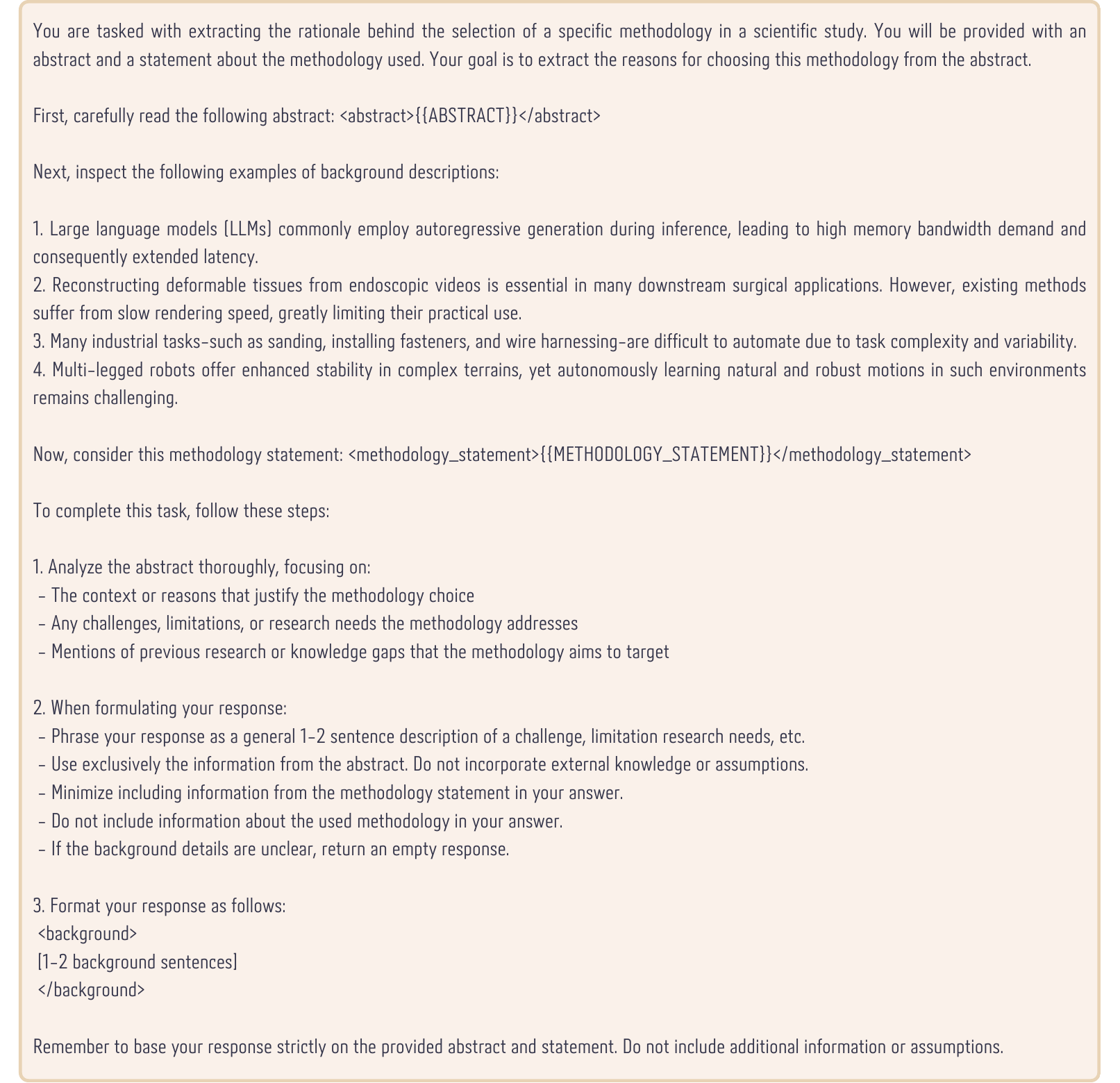}
    \caption{Context extraction prompt. \{\{ABSTRACT\}\} is a placeholder for the input abstract. \{\{METHODOLOGY\_STATEMENT\}\} is a sentence describing the recombination. We build it by filling one of the following templates with the extracted recombination entities: "\textit{Combine <source-entity> and <target-entity>}" for blends and "\textit{Take inspiration from <source-entity> and apply it to <target-entity>}" for inspirations.}
    \label{fig:context_extraction_prompt}
\end{figure*}

\begin{figure}[!htb]
    \centering
    \includegraphics[width=\columnwidth]{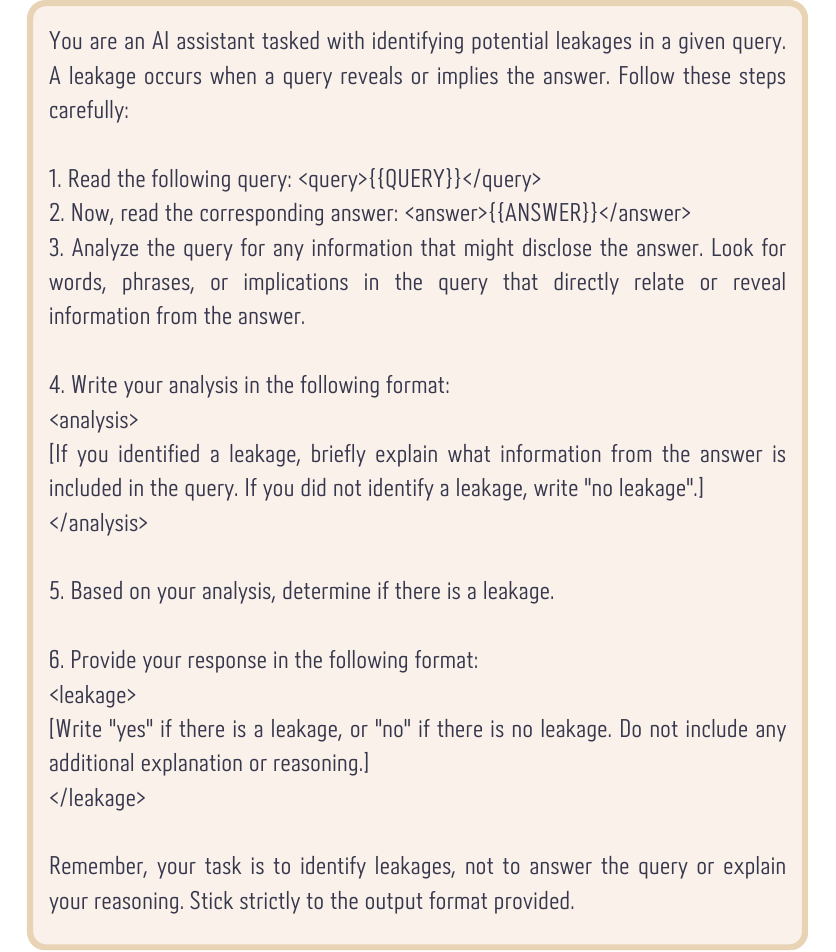}
    \caption{Leak detection prompt.}
    \label{fig:leakage_detect_prompt}
\end{figure}

\paragraph{Context extraction and leakage filtering} We use GPT-4o-mini to extract a few sentences from each abstract describing the background or motivation of the authors using recombination (See prompt on Figure \ref{fig:context_extraction_prompt}). Adding these contexts to the queries helps them be more specific and limits the search space. However, this might introduce leaks into the queries - cases where the extracted context reveals the answer. Table \ref{table:leak_examples} presents leak examples. We utilize GPT-4o-mini again to filter out such cases from the data, using the prompt shown in Figure \ref{fig:leakage_detect_prompt}. In a qualitative analysis of 50 randomly sampled query-answer pairs, we find that a human annotator agrees with 87\% of the model's predictions (whether there is a leak).
Finally, we divide the remaining query-answer pairs into splits as described in Table \ref{table:pred_data_splits} is Section \ref{subsection:recomb_pred}.

\subsection{Prediction baselines}\label{sec:prediction_baselines}
\begin{figure}[!htbp]
    \centering
    \includegraphics[width=\columnwidth]{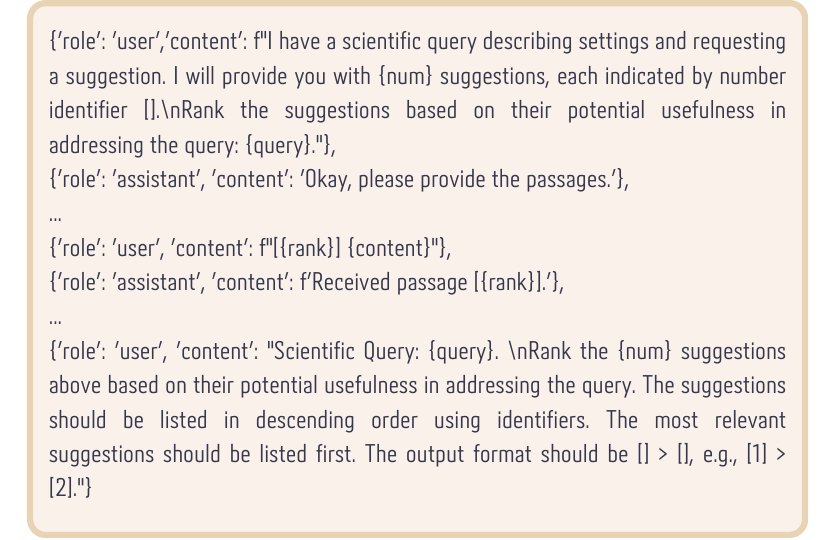}
    \caption{Adjusted RankGPT prompt.}
    \label{fig:rankgpt_prompt}
\end{figure}

We use a bi-encoder architecture for recombination prediction and experiment with three popular encoders as backbones: \texttt{all-mpnet-base-v2} (109M parameters), \texttt{bge-large-en-v1.5} \cite{bge_embedding} (335M parameters) and \texttt{e5-large-v2} \cite{wang2022text} (335M parameters). These models' checkpoints predate 2024, meaning they are unfamiliar with our test set. The model receives a query string composed of a context description, a graph entity, and a relation type and returns a ranked list of answers (other graph nodes). We perform HPO (random grid search of 10 trails) to select the number of training epochs, warmup ratio and learning rate for each model. We use contrastive loss and generate 30 negatives per positive example. Following the literature standard \cite{Teach2020YouCT}, we report metrics in the filtered settings to avoid false negatives. Given the difficulty of the task we focus on ranking only the $12751$ test set entities.
A full summary of our data splits is available on \ref{table:pred_data_splits}. The examples we use to train and evaluate our prediction models contain all collected nodes, including those classified as belonging to the "other" domain.

We utilize RankGPT \cite{Sun2023IsCG} as a strong reranker and apply it to rerank the top-20 predicted results. We employ RankGPT with GPT-4o, a window size of 10 and a step size of 5. Note the information cutoff of GPT-4o is October 2023 \footnote{As stated in https://platform.openai.com/docs/models/gpt-4o}, meaning it is unfamiliar with our test set as well. We use the implementation available in \footnote{https://github.com/sunnweiwei/RankGPT/tree/main}. However, we find that adjusting the default prompt works better for our task. Figure \ref{fig:rankgpt_prompt} shows the modified reranking prompt. The cost of applying the reranker to our data was 60\$.

\subsection{Reranker error analysis}\label{sec:reranker_errors}

\begin{table*}[!htbp]
\centering
\footnotesize
\begin{tabular}{p{7.5cm}p{7.5cm}}
    \toprule
    \multicolumn{2}{c}{\textbf{Reranking error examples}} \\
    \toprule
    \multicolumn{2}{c}{\textit{(i) Multiple plausible answers}} \\
    \midrule
    \multicolumn{2}{p{15cm}}{
    \textbf{Query}: "Traditional reasoning methods in language models often rely on historical information and employ a uni-directional reasoning strategy... This leads to suboptimal decision-making... What would be a good source of inspiration for \textit{enhancing the decision rationality of language models}?"
    } \\
    \midrule
    \textit{\underline{Pre-reranking (top-20)}}\newline
    1. the inherent human attribute of engaging in logical reasoning to facilitate decision-making \newline
2. \marker{principles of rational decision-making} \newline
3. \bmarker{the Level-K framework from game theory and behavioral economics, which extends reasoning from simple reactions to structured strategic depth} \newline
...
    & 
    \textit{\underline{Post-reranking (top-20)}}\newline
1. \bmarker{the Level-K framework from game theory and behavioral economics, which extends reasoning from simple reactions to structured strategic depth} \newline
2. Bayesian inference: conditioning a prior on evidence\newline
...\newline
6. \marker{principles of rational decision-making}\\

    \midrule
    \multicolumn{2}{p{15cm}}{
    \textbf{Query}: "...while Large Language Models (LLMs) excel in various NLP tasks, their ability to generate comprehensive data stories remains underexplored...
What would be a good source of inspiration for \textit{Data-driven storytelling}?"
    } \\
    \midrule
    \textit{\underline{Pre-reranking (top-20)}}\newline
    1. \marker{the human storytelling process} \newline
2. story writing \newline
3. Interactive digital stories \newline
... \newline
9. \bmarker{narrative structure designs} \newline
...
    & 
    \textit{\underline{Post-reranking (top-20)}}\newline
1. story analysis and generation systems\newline
2. generative artificial intelligence (Gen-AI)-driven narrative personalization\newline
3. \bmarker{narrative structure designs} \newline
4. \marker{the human storytelling process} \newline
...\\

    \midrule
    \multicolumn{2}{c}{\textit{(ii) Semantically similar variants}} \\
    \midrule
        \multicolumn{2}{p{15cm}}{
    \textbf{Query}: "Prior methods for aligning large language models face challenges in tuning to maximize non-differentiable and non-binary objectives...This highlights a need for a more flexible approach that can generalize to various user preferences... while maintaining alignment... What could we blend with \textit{reinforcement learning via human feedback} to address the described settings?"}\\
    \midrule
    \textit{\underline{Pre-reranking (top-20)}}\newline
    1. aligning Large Language Models with human preferences \newline
    2. Direct Preference Optimization for preference alignment \newline
    3. \marker{direct preference optimization} \newline
    ... \newline
    5. \bmarker{State-of-the-art language model fine-tuning techniques, such as Direct Preference Optimization} \newline
    ...
    & 
    
    \textit{\underline{Post-reranking (top-20)}}\newline
    1. Direct Preference Optimization for preference alignment\newline
2. \bmarker{State-of-the-art language model fine-tuning techniques, such as Direct Preference Optimization}\newline
3. contrastive learning-based methods like Direct Preference Optimization\newline
4. a Semi-Policy Preference Optimization method\newline
5. \marker{direct preference optimization}\newline
...\\
    \bottomrule
\end{tabular}
\caption{\revision{Illustrative examples where the reranker preferred a \bmarker{different answer} over the \marker{gold one}.}}\label{tab:ranking_errors}
\end{table*}

\revision{In Section~\ref{subsec2:recombination_pred_results}, we show that reranking the top-$20$ answers retrieved by our best-performing prediction model (all-mpnet-base-v2$_\text{finetuned}$) can sometimes \textbf{lower} the rank of the gold candidate. To better understand the underlying causes of such reranking failures, we conduct an error analysis of $30$ representative cases. Our goal in this section is to describe common patterns in these errors and highlight particularly challenging scenarios that may inform future progress.}

\paragraph{(i) Multiple plausible answers.}
\revision{In some cases, the reranker correctly identifies a strong and highly relevant candidate, and ranks it above the gold even though both answers are valid. These errors stem not from a lack of understanding, but from the presence of several equally reasonable responses. For instance, in Table~\ref{tab:ranking_errors} (top), the reranker promotes a conceptually grounded strategy from game theory over a more generic gold response about rational decision principles.}

\paragraph{(ii) Semantically similar variants.}
\revision{Another common error involves the reranker prioritizing paraphrased or reformulated versions of the gold answer. While these candidates are semantically close to the gold, the gold itself may fall in rank due to redundancy. As shown in Table~\ref{tab:ranking_errors} (bottom), several variants of "Direct Preference Optimization" receive high rankings, but the original mention of the method is pushed downward, possibly due to lexical overlap penalties or insufficient canonicalization.}

\revision{These examples highlight nuanced challenges in reranking systems, such as handling redundancy and conceptual equivalence.}

\section{User study additional details}\label{app:user_study}

\begin{figure}[!htbp]
    \centering
    \includegraphics[width=\columnwidth]{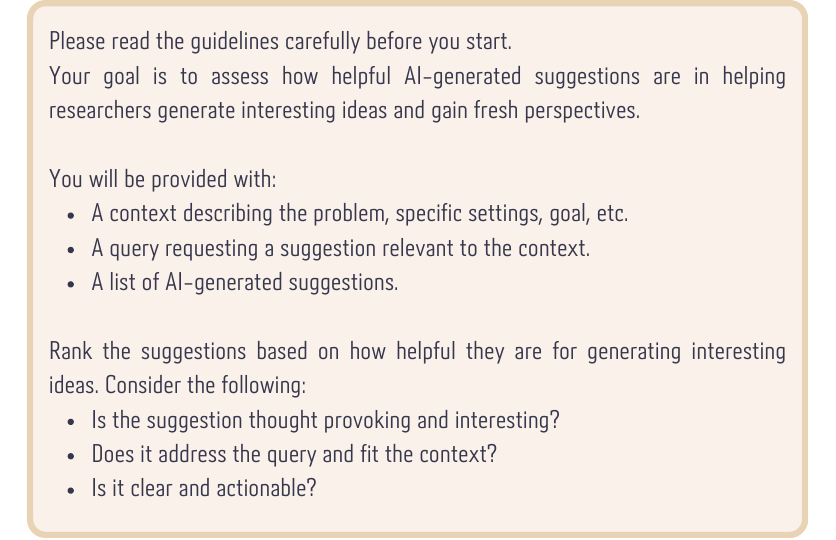}
    \caption{User study guidelines.}
    \label{fig:study-guidelines}
\end{figure}

\begin{figure}[!htbp]
    \centering
    \includegraphics[width=\columnwidth]{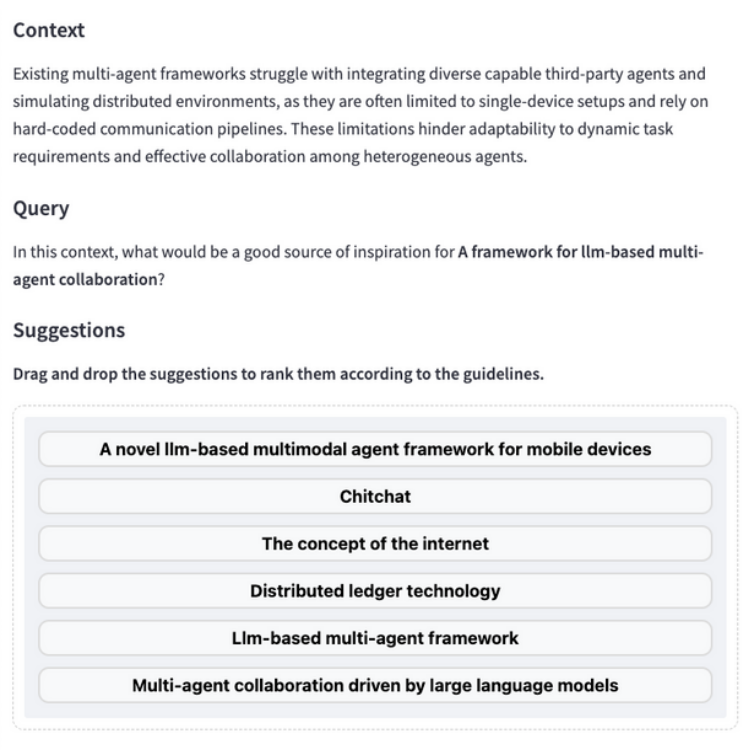}
    \caption{User study interface.}
    \label{fig:study-interface}
\end{figure}

We request each to fill out a form asking in what scientific domains they feel comfortable reading papers and a short description of their research area. We then used \texttt{granite-embedding-125m-english} to retrieve semantically similar contexts to this description from the relevant arXiv categories. We manually verify that the retrieved contexts match the description and discard examples with poorly extracted information (e.g., the context begins with "\textit{This study reviews the problem of...}" instead of directly describing the source study problem). In addition, we let the volunteers mark an example as "ill-defined", in which case we ignore their inputs.
We conduct a 10-minute training session with each volunteer, requesting them to read the instructions and explain the task. Figure \ref{fig:study-guidelines} presents the instructions given to the participants in the study. Figure \ref{fig:study-interface} presents the web interface of the annotation platform.

\subsection{Predictions examples}\label{sec:pred_examples}
\revision{Table~\ref{tab:pred_examples} shows a selection of model predictions that participants rated as most helpful for inspiring research directions. These examples highlight how CHIMERA-trained models can move beyond surface-level associations to propose insightful cross-domain inspirations, for instance, linking harmful meme detection to visual commonsense reasoning, or drawing on neuroscience to improve LLM knowledge retention. Such predictions demonstrate CHIMERA’s potential to power ideation tools that help researchers identify novel, actionable directions for future work.}

\begin{table*}[!htbp]
    \centering
    \footnotesize
    \begin{tabular}{p{15cm}}
    \toprule
         \multicolumn{1}{c}{\textbf{Prediction Examples}}\\ \midrule
         \textbf{Query}: Existing large language models (LLMs) underperform in legal judgment prediction due to challenges in understanding case complexities and distinguishing between similar charges. This highlights a need for improved methodologies that can effectively address these issues to enhance judicial efficiency.
         In this context, what would be a good source of inspiration for \textit{legal judgment prediction}? \newline\newline
         \textbf{Prediction}: The Issue, Rule, Application, Conclusion (Issue, Rule, Application, Conclusion) framework, a sequential reasoning approach used by lawyers. \\ \midrule

        \textbf{Query}: The exponential increase in scientific literature makes it challenging for researchers to stay current with recent advances and identify meaningful research directions. Existing methods for idea generation either trivially prompt large language models or directly expose them to extensive literature without indicating useful information, highlighting a need for more effective approaches to research ideation.
         In this context, what would be a good source of inspiration for \textit{automating the generation of novel research ideas}? \newline\newline
         \textbf{Prediction}: Emulate the human process of inducing collective creativity through engaging discussions with participants from diverse backgrounds and perspectives. \\ \midrule

        \textbf{Query}:The challenge of detecting harmful memes lies in the implicit meanings embedded within them, which are not explicitly conveyed through surface text and images. Existing harmful meme detection methods fail to provide readable explanations that reveal these implicit meanings, highlighting a significant gap in the ability to support detection decisions.
         In this context, what would be a good source of inspiration for \textit{detecting harmful memes}? \newline\newline
         \textbf{Prediction}: Visual commonsense discovery in computer vision. \\ \midrule

                 \textbf{Query}: Large language models (LLMs) often struggle to provide up-to-date information due to their one-time training and the constantly evolving nature of the world. Existing approaches to keep LLMs current face difficulties in extracting stored knowledge, highlighting a need for improved methods of knowledge acquisition from raw documents. In this context, what would be a good source of inspiration for \textit{improving an llm's ability to effectively acquire new knowledge from raw documents}? \newline\newline
         \textbf{Prediction}: Neuroscience, where the human brain often sheds outdated information to improve the retention of crucial knowledge and facilitate the acquisition of new information. \\ 
         
    \bottomrule
    \end{tabular}
    \caption{\revision{Examples of recombination directions predicted by our model and rated as most inspiring by user study participants. Each prediction links a scientific challenge with a cross-domain concept, illustrating \kbname’s potential to support creative research ideation.}}\label{tab:pred_examples}
\end{table*}

\section{Comparison to other information extraction methods}\label{sec:extraction_comp}
\begin{figure*}[!htbp]
    \centering
    \begin{subfigure}[b]{\textwidth}
        \centering
        \includegraphics[width=0.9\textwidth]{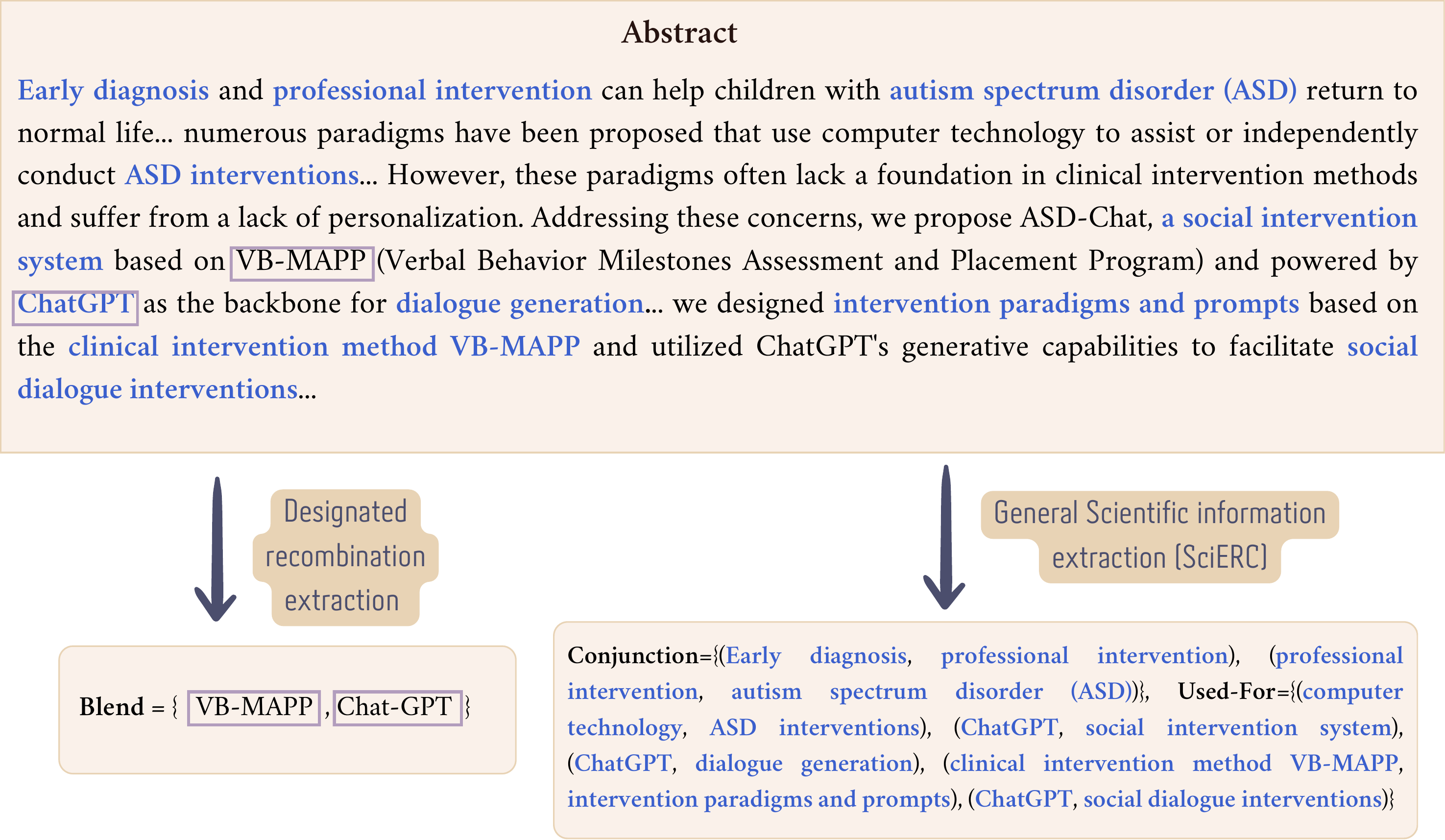}
        \caption{Comparison to recombination extraction using a general scientific IE schema (SciERC)}\label{fig:scierc_schema}
    \end{subfigure}
    
    \vspace{2em}
    
    \begin{subfigure}[b]{\textwidth}
        \centering
        \includegraphics[width=0.9\textwidth]{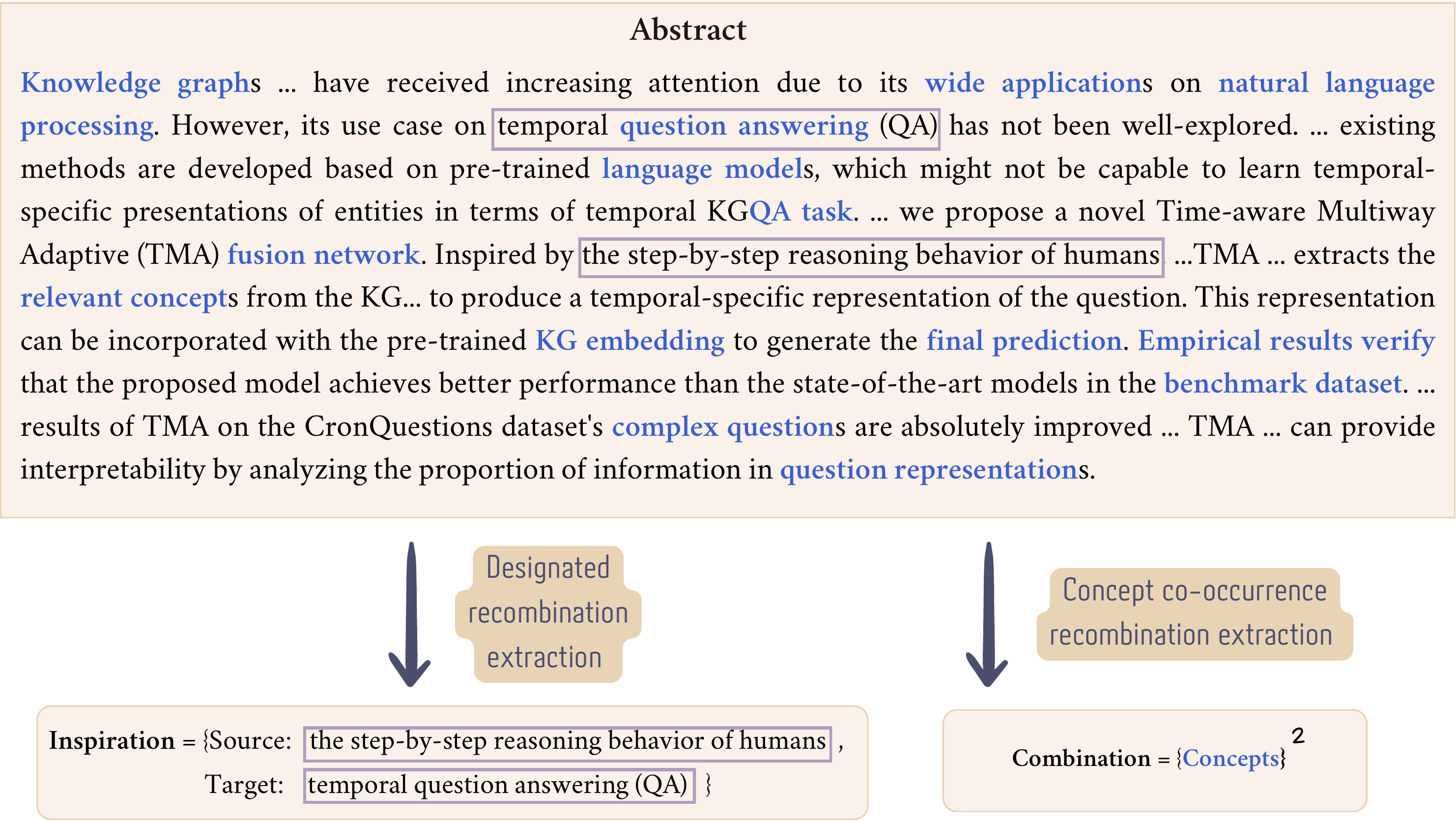}
        \caption{Comparison to recombination extraction using concept co-occurrence.}\label{fig:concept_co_occurrence}
    \end{subfigure}
    \caption{Comparison of our designate recombination extraction method to alternative approaches.  Figure \ref{fig:scierc_schema}: General recombination extraction schemas lack fitting relation types to capture recombinations, which results in capturing plenty of irrelevant relations ("\textit{Early diagnosis}" $\longleftrightarrow$ "\textit{professional intervention}"). Figure \ref{fig:concept_co_occurrence}: Recombination extraction using concept co-occurrence might be nonsensical ("\textit{wide application}" $\longleftrightarrow$ "\textit{final prediction}") or even misleading ("\textit{question answering}" $\longleftrightarrow$ "\textit{language models}")). }
    \label{fig:extraction_comparison}
\end{figure*}

Both general scientific extraction and concept co-occurrence struggle to capture concise and accurate recombination relations, as can be seen in Figure \ref{fig:extraction_comparison}. Figure \ref{fig:scierc_schema} presents how general scientific IE schemas lack relation types to model recombinations. The figure presents the results of our specialized extraction method besides a transformer-based extraction model \cite{Hennen2024ITERIT} finetuned on SciERC \cite{Luan2018MultiTaskIO}, a general IE schema. While our new data schema easily models the recombinant connection between two techniques: "\textit{BV-MAPP (Verbal Behavior Milestones Assessment and Placement Program)}", "\textit{ChatGPT}" as a concept \textit{blend}, the SciERC extraction schema isn't equipped with proper relation types for this. As a result, it captures mostly irrelevant information for our task (e.g background details as "\textit{Early diagnosis}" or "\textit{professional intervention}"). Figure \ref{fig:concept_co_occurrence} shows how recombination extraction using concept co-occurrence might be misleading. In this method, each pair of canonical scientific concepts (e.g, \textit{neural networks}) that co-occur within the same abstract are considered a recombination. The figure presents an example of using AI-related concepts curated by \citet{Krenn2022ForecastingTF} for recombination extraction, alongside recombination extracted using our designated approach. Note that when using concept co-occurrence, the extracted recombinations are essentially $\{concepts\}^2$, which might be imprecise, and capture meaningless recombinations (e.g., "\textit{wide application}" recombined with "\textit{final prediction}") or misleading recombinations (e.g., "\textit{question answering}" with "\textit{language models}", which explicitly presented by the authors as a lacking approach for the task). In comparison, our new extraction schema neatly models the main recombiant relation presented in the text as taking inspiration from "\textit{the step-by-step reasoning behavior of humans}" for "\textit{temporal question answering}."

\end{document}